\pdfoutput=1
\documentclass[11pt]{article}

\usepackage{ACL2023}

\usepackage{times}
\usepackage{amsmath}
\usepackage{amssymb}
\usepackage{pifont}
\usepackage{latexsym}
\newcommand{\cmark}{\ding{51}}%
\newcommand{\xmark}{\ding{55}}%

\usepackage[T1]{fontenc}

\usepackage[utf8]{inputenc}

\usepackage{inconsolata}
\usepackage{booktabs}
\usepackage{float}
\usepackage{hyperref}
\usepackage{graphicx}
\graphicspath{ {./images/} }
\usepackage{multirow}
\usepackage{ragged2e}
\usepackage{color}
\usepackage{array}
\usepackage{arydshln}
\usepackage{subcaption}
\usepackage{linguex}
\usepackage[switch]{lineno}
\usepackage{tcolorbox}
\usepackage{tablefootnote}
\usepackage{colortbl}
\newcommand{\corpname}{GUM-SAGE}

\title{GUM-SAGE: A Novel Dataset and Approach for Graded Entity Salience Prediction}
\author{Jessica Lin \and Amir Zeldes\\
         Department of Linguistics \\ Georgetown University \\ \texttt{\{yl1290, amir.zeldes\}@georgetown.edu}}

\begin{document}
\maketitle
\begin{abstract}
Determining and ranking the most salient entities in a text is critical for user-facing systems, especially as users increasingly rely on models to interpret long documents they only partially read. Graded entity salience addresses this need by assigning entities scores that reflect their relative importance in a text. Existing approaches fall into two main categories: subjective judgments of salience, which allow for gradient scoring but lack consistency, and summarization-based methods, which define salience as mention-worthiness in a summary, promoting explainability but limiting outputs to binary labels (entities are either summary-worthy or not). In this paper, we introduce a novel approach for graded entity salience that combines the strengths of both approaches. Using an English dataset spanning 12 spoken and written genres, we collect 5 summaries per document and calculate each entity's salience score based on its presence across these summaries. Our approach shows stronger correlation with scores based on human summaries and alignments, and outperforms existing techniques, including LLMs. We release our data and code at \url{https://github.com/jl908069/gum_sum_salience}\footnote{Data is also available from \url{https://github.com/amir-zeldes/gum}.} to support further research on graded salient entity extraction. 
\end{abstract}

\section{Introduction}

Salient entity extraction (SEE) is the task of identifying the most central entities mentioned in an arbitrary document in a given language, based on their contribution to the overall meaning of the document \cite{gamon2013identifying}. SEE has a range of applications, for example in news search and analysis, as well as summarization \cite{asgarieh2024scalabledetectionsaliententities}, since users may want to categorize articles based on mentioned salient (but not non-salient) entities, or ensure that summary information focuses on salient, rather than tangential entities. 

Two key challenges inherent in work on SEE are the gradualness and subjectivity involved in human salience judgments. For example, in a biography of Albert Einstein, Einstein is likely to be the most salient person, but other people mentioned can still be more salient, e.g.~Danish physicist Niels Bohr, with whom Einstein entered into prominent debates, or less so, e.g.~Einstein's uncle Jakob -- both are mentioned in Einstein's Wikipedia article, the latter only briefly but the former 11 times. Although human raters are likely to easily agree that Bohr is more salient in Einstein's Wikipedia page than Jakob Einstein, there are many subtle cases on which they disagree: \citet{dojchinovski2016crowdsourced} showed that crowdsourced entity salience annotations contained nearly 20\% of labels that had to be discarded as `untrustworthy', and achieved only around 63\% agreement. 

Other approaches opt to exploit additional properties of documents for a more operationalizable definition of salience, for example based on hyperlinks in the document \cite{wu2020wn} or the mention of entities in a summary of the document \cite{dunietz2014new,lin-zeldes-2024-gumsley}. While these approaches reach much higher agreement (over 97\% in \citealt{lin-zeldes-2024-gumsley}), they are limited to single, binary judgments (an entity either appears in a summary or not, is either hyperlinked or not, etc.) and are more sensitive to variability in the underlying properties (many documents contain no links, a link may or may not be added, the summary could have been different, etc.).

In this paper, we aim to combine the benefits of clearly operationalized approaches to entity salience, specifically in the paradigm of summary-based salience, with the advantages of graded salience labels derived from multiple aggregated sources. Our approach relies on collecting multiple, separate summaries of each document, and aligning mentions to the original text to create numerical salience scores based on the number of summaries mentioning each entity (e.g.~5/5 summaries would mention Einstein in his biography, some might mention Bohr, and probably none would mention his uncle). To the best of our knowledge, this is the first attempt to cast summarization-based salience as a regression, rather than a classification problem.

The main contributions of this paper are:

\begin{itemize}
    \item A novel approach to graded summary-based entity salience prediction, demonstrating a nearly 17-point improvement in F1 score and better correlation with human summary and alignment-based salience scores compared to  leading LLMs
    \item A new dataset based on the openly available UD English GUM corpus \cite{zeldes2017gum}, semi-automatically enriched with 5 aligned summaries and graded salience scores for all named and non-named entities
    \item A thorough evaluation of models and systems used to create our data, including fully manually constructed test and dev sets
    \item Analysis of error patterns in models' entity salience predictions and their correlation with human annotated salience 
\end{itemize}

\section{Related Work}
\begin{table*}[htb]
    \centering
    \renewcommand{\arraystretch}{1.2} 
    \scalebox{0.5}{
    \begin{tabular}{@{}lclrrrrrc@{}}
    \toprule
    Datasets & Multi-Genre & Multi-types & \# of Documents & \# of Entities & \% of Salient Entities & Entity Annotations & Salience Labels & Graded Salience \\ \midrule
    MDA \cite{gamon2013identifying} & \cmark & \xmark & $\approx$ 50,000 & 2,414 & < 5 \% & proprietary NLP pipeline & soft labeling & \cmark \\
    NYT-Salience \cite{dunietz2014new} & \xmark & \xmark & 110,639 & 2,229,728 & $\approx$ 14 \% & proprietary NLP pipeline & automated & \xmark \\
    Reuters-128 Salience \cite{dojchinovski2016crowdsourced} & \xmark & \xmark & 128 & 4,429 & 18\% & manual annotation & crowdsourcing & \cmark \\
    The Wikinews dataset \cite{trani2018sel} & \xmark & \xmark & 365 & $\approx$ 4,400 & $\approx$ 10 \% & manual annotation & crowdsourcing & \cmark \\
    WN-Salience \cite{wu2020wn} & \xmark & \xmark & 6,968 & 888 & 16\% & manual annotation & automatic derivation & \xmark \\
    EntSUM \cite{maddela-etal-2022-entsum} & \xmark & \xmark & 693 & 7,854 & 39\% & semi-automated & crowdsourcing &  \cmark\\
    WikiQA-Salience \cite{bullough-etal-2024-predicting} & \xmark & \xmark & 687 Q/A pairs & 2,113 & $\approx$ 52\% high salience & open source NLP library & crowdsourcing & \cmark\\ 
    CIS Entity Salience \cite{sekulic2024towards} & \xmark & \xmark & 120 Q/A pairs & $\sim$400 & $\approx$ 63\% & open source NLP library & crowdsourcing & \cmark\\
    GUMsley \cite{lin-zeldes-2024-gumsley} & \cmark & \cmark & 213 & 29,899 & 7\% & manual annotation & semi-automated & \xmark \\
    \bottomrule
    \end{tabular}}
    \caption{Statistics of existing entity salience datasets. The column \texttt{Multi-types} shows whether the dataset covers diverse types of entities (named entities, non-named entities, wiki-linked entities) and NPs (e.g., verbal NPs).}
    \label{tab:datasets_diff_general}
\end{table*}

The increasing importance of SEE is reflected in the expanding number of annotated datasets, employing different strategies for entity recognition and salience labeling (see Table \ref{tab:datasets_diff_general}). However, building a reliable dataset with consistent entity salience annotations remains a significant challenge for a number of reasons, including lack of reliable and exhaustive entity annotations, the absence of consistent guidelines for entity salience, subjectivity in the assignment of salience scores, and limited data availability across text genres, with previous work focusing almost only on news and Wikipedia material. \par

To ensure the reliability of an entity salience dataset, the first step is to adopt a robust method for identifying entities within a document. Some work \cite{dunietz2014new, gamon2013identifying} has utilized multi-step automatic pipelines (including NP extraction, coreference resolution, and a named entity recognizer) to identify entities, while others \cite{dojchinovski2016crowdsourced,trani2018sel,wu2020wn} have undertaken manual annotation. The latter is potentially more accurate but expensive, while NLP pipelines are cheap but may propagate errors to later steps. Furthermore, salient entities in a document are not necessarily named entities, but also non-named ones. Most previous datasets \cite{bullough-etal-2024-predicting, dojchinovski2016crowdsourced, dunietz2014new, gamon2013identifying, maddela-etal-2022-entsum, sekulic2024towards, trani2018sel, wu2020wn}
include only named entities, leaving out common noun entities that may be salient to humans -- in fact, some documents contain no named entities, but we would still assume some of the non-named entities will be more salient than others. \par

The second challenge in building a reliable entity salience dataset is how to minimize noise in salience labels and apply consistent guidelines. Entity salience labels have been derived either through crowdsourcing, gathering ratings from multiple non-expert raters to determine salient entities \cite{bullough-etal-2024-predicting, dojchinovski2016crowdsourced,maddela-etal-2022-entsum, sekulic2024towards, trani2018sel}, or by employing proxy methods such as abstracts or writer-assigned Wikinews categories \cite{gamon2013identifying,dunietz2014new,wu2020wn}.
While crowdsourcing can surpass automated methods in performance, it is inherently noisy and prone to bias, as opinions on what is salient can be unpredictable \cite{maddela-etal-2022-entsum}. On the other hand, using proxies has been shown to be less noisy (i.e.~more reproducible), but can suffer from a lack of reliability in the proxies themselves. For example, the NYT (\textbf{N}ew \textbf{Y}ork \textbf{T}imes)-salience dataset \cite{dunietz2014new} relied on found news abstracts to identify salient entities. This approach has several limitations: First, the derived salience labels may be less reliable due to the lack of clear and consistent guidelines for summaries (e.g.~length, style). Second, relying on news article abstracts restricts the dataset to certain types of genres (i.e.~news). \par

In this study, we adopt a regimented approach similar to the NYT salience corpus (\citealt{dunietz2014new}), which identifies salient entities using summaries. To improve annotation reliability, we crowdsourced summaries across 12 English genres, following very specific guidelines proposed by \citet{LiuZeldes2023}. This method increases the reliability of the summaries as proxies for salience, and in turn the consistency of salient entity annotations, compared to reliance on found abstracts limited to news articles, which were not designed for this task. Additionally, it allows us to obtain graded salience judgments without sacrificing this reliability.

\section{Dataset}
Our dataset, called \corpname{} (\textbf{GUM}-based \textbf{S}ummary \textbf{A}ligned \textbf{G}raded \textbf{E}ntities) is based on the GUM corpus \cite{zeldes2017gum}, an open-access manually annotated, multilayer resource for English. The corpus spans over 200K tokens across 12 different text genres (see Table~\ref{tb:stats}), and includes Universal Dependencies (UD) parses \cite{de-marneffe-etal-2021-universal}, detailed entity annotations such as entity types and Wikification links \cite{lin-zeldes-2021-wikigum}, coreference resolution \cite{zhu-etal-2021-ontogum}, and discourse parses \cite{liu-zeldes-2023-cant}. Additionally, the dataset provides an expert-written summary for each document (\citealt{LiuZeldes2023}), aligned to the entity annotations for entities mentioned in the summary \cite{lin-zeldes-2024-gumsley}, which we leverage for evaluation below. \par

In this paper we add entity salience scores (0-5) for all named and non-named entities in the data using an SEE pipeline with two components: Summary Crowdsourcing \& Generation (Section \ref{sec: sum-crowdsourcing-generation}) and Entity Alignment (Section \ref{sec: auto-alignment}). We assume that it is difficult to summarize a text without mentioning its most salient entities, and that salient entities will therefore tend to appear in summaries, while spuriously mentioned entities will not recur in many summaries. The SEE pipeline therefore collects multiple summaries per document, aligning mentions to assign salience scores based on the number of summaries mentioning the entity. We evaluate the accuracy of our approach in Section \ref{sec: alignment-eval}.\par

\begin{table*}[htbp]
\centering
\resizebox{\textwidth}{!}{%
\renewcommand{\arraystretch}{1.1} 
\begin{tabular}{llrrrrrrrrrrrrr}
\toprule
\textbf{} & \textbf{} & \textbf{academic} & \textbf{bio} & \textbf{conversation} & \textbf{fiction} & \textbf{interview} & \textbf{news} & \textbf{reddit} & \textbf{speech} & \textbf{textbook} & \textbf{vlog} & \textbf{voyage} & \textbf{wikihow} & \textbf{TOTAL} \\ \midrule
\textbf{Documents} &  & 18 & 20 & 14 & 19 & 19 & 23 & 18 & 15 & 15 & 15 & 18 & 19 & \textbf{213} \\ 
\textbf{Tokens} &  & 17,905 & 18,554 & 14,307 & 18,003 & 16,504 & 21,767 & 17,986 & 13,195 & 14,451 & 14,784 & 17,984 & 18,341 & \textbf{203,781} \\ 
\textbf{Mentions} &  & 5,045 & 5,768 & 4,094 & 4,974 & 5,211 & 4,720 & 4,544 & 4,847 & 4,719 & 4,499 & 4,471 & 4,468 & \textbf{57,360} \\ 
\textbf{Entities} &  & 3,251 & 3,324 & 1,363 & 2,352 & 2,642 & 2,579 & 2,364 & 2,573 & 2,885 & 1,626 & 2,957 & 2,384 & \textbf{32,300} \\ 
\textbf{Avg \# of Entities} &  & 181 & 166 & 97 & 124 & 139 & 112 & 131 & 172 & 192 & 108 & 164 & 125 & \textbf{148} \\ 
\textbf{\% Salient Entities} &  & 6.3 & 9.1 & 9.4 & 8.0 & 9.6 & 11.6 & 12.2 & 14.4 & 15.8 & 16.8 & 19.4 & 32.9 & \textbf{13.8} \\ 
\textbf{\% of Top1 Salient Entities} &  & 0.9 & 1.2 & 3.2 & 1.5 & 2.0 & 2.3 & 1.5 & 2.1 & 1.8 & 2.4 & 2.5 & 3.6 & \textbf{2.1} \\ 
\textbf{\% of Top3 Salient Entities} &  & 1.9 & 2.8 & 6.6 & 3.5 & 3.7 & 4.6 & 4.7 & 4.6 & 4.9 & 6.5 & 5.1 & 9.9 & \textbf{4.9} \\ \bottomrule
\end{tabular}}
\caption{Overview of \corpname{}. Top1 salient entities are those with a score of 5; Top3 refers to entities with scores of 3, 4, or 5. \% salient entities = number of all salient entities (score 1-5) / total number of entities. Avg entities per summary = \# of entities / \# of documents in the genre.}
\label{tb:stats}
\end{table*}

\subsection{Summary Crowdsourcing \& Generation}
\label{sec: sum-crowdsourcing-generation}
Each document in GUM is already accompanied by a single expert-written summary, and an additional second human-written summary is provided for each of the 24 test documents \cite{LiuZeldes2023}. However because our approach to salience is based heavily on summary content, which can vary, a single summary may be inadequate to identify salient entities within a document, by either missing some salient entities, or containing spurious ones. We therefore crowdsource or generate summaries for our data.


\paragraph{Summary Crowdsourcing}
In the summary crowdsourcing task, each annotator is asked to read eight documents from different genres before writing a one-sentence summary for the document, which should `substitute reading the text', focus on `who did what to whom', and, space allowing, `when, where and how', but may not exceed 380 characters, following \citet{LiuZeldes2023}. Annotators were also instructed not to mention anything not mentioned in the document, and to adhere as closely as possible to the document's vocabulary and phrasing. All of the crowdsourced summaries were manually checked by one of the authors to ensure that they follow the guidelines. Most of the summaries did so, though a small portion deviated in two key areas: (i) Mentioning facts not mentioned in the text, such as speaker names not explicitly stated but identifiable from context; (ii) Using shell nouns like ``this reddit post'', which should generally be avoided if they are not unambiguously identifiable in the text (e.g.~a writer states this is a reddit post). Any summaries that did not comply with guidelines were minimally manually corrected to maintain consistency, without otherwise altering their meaning.\par

\paragraph{Summary Generation}
We select four recent LLMs -- \texttt{GPT-4o \cite{openai2024gpt4}}, \texttt{Claude 3.5 Sonnet} \cite{anthropic2024claude3}, \texttt{Llama 3.2 3B Instruct} \cite{meta2024llama3}, and \texttt{Qwen 2.5 7B Instruct} \cite{qwen2.5} -- to create four ``silver'' summaries\footnote{See Section~\ref{sec:appendix-auto-sumeval} for details on the quantitative evaluation of generated summaries.} for each of the 165 training set documents, matching the length and style of GUM summaries. Along with one gold summary per document, this resulted in five summaries per document. All models were instructed to produce a one-sentence summary and were given examples from the dev set for the genre in question\footnote{See Appendix~\ref{sec:appendix-sg} for prompt details.}.\par
Summaries violating the length limit were resolved by re-prompting the LLM (in the same session) to abbreviate to the required length. Finally, minimal automatic corrections were applied, such as replacing periods with semicolons if models outputted more than one sentence, in order to ensure compatibility with the manual gold dev and test sets of the \corpname{}.

\subsection{Entity Alignment}
\label{sec: auto-alignment}
We aligned mentions in human-written and system-generated summaries with those in the document using several methods, from rule-based methods to NLP pipelines, to prompt-based LLMs, as well as manual alignment for the dev/test data\footnote{See Appendix~\ref{sec:appendix-interface} for details on the interface we used.}. \par

\begin{itemize}
    \item \textbf{String Match}: To achieve high precision in aligning mentions to summaries with corresponding mentions, 
    we use exact and partial string matching, iterating over the gold entity annotations in each GUM document: 
    for multi-word mentions (>2 tokens), we allow for partial matching when more than 3 contained tokens appear exactly in the summary, excluding stop words. For example, if a summary mentions `the prevalence of racial discrimination in the United States', and the document mentions `The prevalence of discrimination across racial groups in contemporary America', the module considers the entity to appear in the summary due to a substring match (\textit{prevalence, racial, discrimination}). This approach provides flexibility in cases where the phrasing of mentions between the summary and the document may differ. 
    \item \textbf{Stanza Coreference Model} \cite{liu-etal-2024-mscaw}: We concatenate each summary to its document and use the Stanza coreference model, trained on CorefUD \cite{nedoluzhko-etal-2022-corefud} with \texttt{XLM-RoBERTa-large} \cite{conneau-etal-2020-unsupervised}, to align mentions between documents and their summaries. Each mention in the document predicted to be in a coreference cluster with any mention in the summary is considered to be included in the summary, and therefore salient.
    \item \textbf{GPT-4o} \cite{openai2024gpt4}: We used\footnote{See Appendix~\ref{sec:appendix-eval} for prompt details.} GPT-4o to determine whether each entity in the document is also mentioned in the corresponding summary, prompting it to take synonyms and alternative phrases into account. We expect GPT-4o to achieve high precision, given its capability to perform at human-level accuracy on various benchmarks. However in initial experiments, feeding the model all document entities at once harmed recall, with the model struggling to identify all relevant matches in a large batch. To optimize recall, we feed the model batches of 15-20 document entities at a time, asking it to match these with those in the summary. This process is repeated through multiple non-overlapping queries for each document, ensuring all entities are evaluated without overwhelming the model. 
    Since we elicit binary judgments per entity, the model's responses can easily be parsed to extract an alignment, which we aggregate across queries. 
    \item \textbf{Ensemble Learning}: To enhance alignment accuracy, we train a logistic regression model on the manually corrected dev set to predict entity salience. The model uses binary predictions from the String Match module, Stanza, and GPT4o as features, and includes other features available from GUM annotations, such as entity type (person, organization, etc.), genre, and document position, achieving stronger performance than any individual module (Table~\ref{tb:alignment_score}). The position feature is defined as the entity's order in the document, e.g.~the first entity has a position of 1, etc.
\end{itemize}

\section{Evaluating Alignment}
\label{sec: alignment-eval}
Table~\ref{tb:alignment_score} presents alignment scores for different alignment methods in Section~\ref{sec: auto-alignment} in matching entities shared between documents and summaries. Note that reported scores refer to the test partition only, since we use the dev partition as training data for the ensemble.\footnote{The alignment for both the test and dev partitions was manually corrected for all summaries using the interface described in Appendix~\ref{sec:appendix-interface}.}\par

\begin{table}[h!]
\centering
\scalebox{0.7}{
\begin{tabular}{p{3.8cm}cccccc}
\toprule
\textbf{Alignment component} & \multicolumn{3}{c}{\textbf{Micro Average}} & \multicolumn{3}{c}{\textbf{Macro Average}} \\
\cmidrule(lr){2-4} \cmidrule(lr){5-7}
 & \textbf{P} & \textbf{R} & \textbf{F} & \textbf{P} & \textbf{R} & \textbf{F} \\
\toprule
Flan-T5-XL & 0.95 & 0.08 & 0.14 & 0.57 & 0.09 & 0.15 \\
\midrule
String match & 0.98 & 0.39 & 0.56 & 0.97 & 0.37 & 0.52 \\
\midrule
GPT-4o (avg 3 runs) & 0.80 & 0.71 & 0.75 & 0.82 & 0.72 & 0.76 \\
\midrule
Stanza Coref & 0.73 & 0.82 & 0.77 & 0.74 & 0.82 & 0.77 \\
\midrule
Ensemble learning & 0.98 & 0.98 & \textbf{0.98} & 0.98 & 0.92 & \textbf{0.95} \\
\midrule
Ensemble learning (for the positive class) & \multicolumn{6}{c}{0.98 / 0.83 / 0.90} \\
\bottomrule
\end{tabular}}
\caption{Alignment scores for different alignment components on the test set of \corpname{}. The best performing F1 scores are in \textbf{bold}.}
\label{tb:alignment_score}
\end{table}

\begin{figure}[htb]
\centering
\includegraphics[width=0.5\textwidth, trim={0pt 100pt 0pt 100pt}, clip]{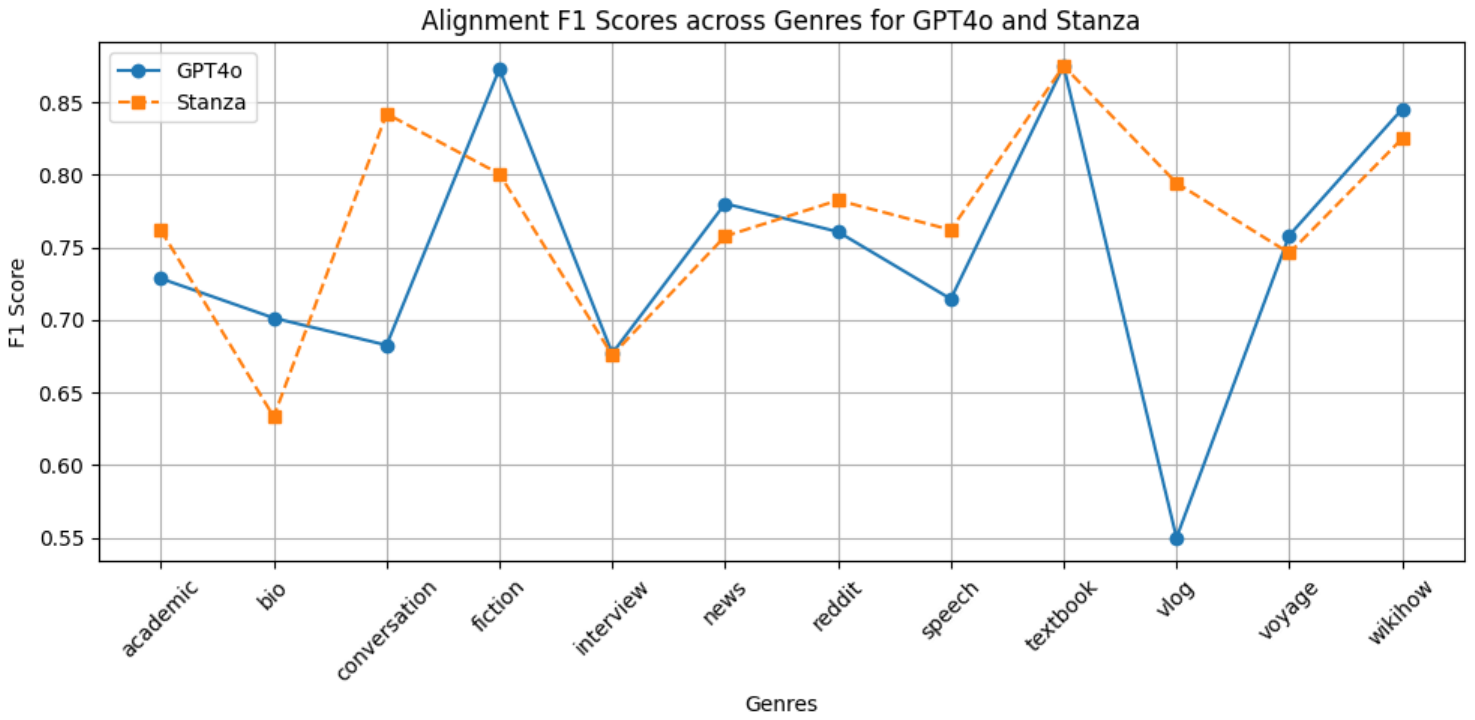}
\caption{Alignment F1 Scores on the test set across genres for GPT4o and Stanza.}
\label{fig:align_f1}
\end{figure}

Interestingly, the Stanza coreference model achieves marginally better overall performance compared to GPT-4o, particularly in \textit{conversation} and \textit{vlog} genres (Figure~\ref{fig:align_f1}), despite GPT-4o's vastly larger training data and parameter count, and both models having access to the same inputs (the document, including gold speaker labels for spoken data). We speculate that Stanza's dedicated coreference architecture benefits from optimization for mention detection, a critical task where prompt-based LLMs like GPT-4o struggle, as noted by \citet{sundar2024major} and \citet{le2023large}, perhaps due to the less natural task of exhaustively identifying all mentions, including nested ones.
Conversational genres like unscripted conversations and vlogs, which involve frequent pronoun use, informal language, and rapid speaker shifts seem to be particularly challenging. Secondly, Stanza benefits from its training on CorefUD \cite{nedoluzhko-etal-2022-corefud}, which includes GUM as the largest open English coreference dataset, making the model familiar with GUM's coreference span definitions (see \citealt{Zeldes2022}), and allowing it to excel in the genres included in the corpus (however note that the Stanza model could not have been exposed to our summaries, which are novel, and is not trained on the test data).

Although no single module achieves satisfactory performance for creating a new benchmark dataset — a key goal of this paper — the ensemble learning approach in Table~\ref{tb:alignment_score} stands out by leveraging the distinct strengths of each component method. While string matching and GPT-4o achieve high precision by identifying clear entity matches, Stanza's coreference architecture provides higher recall by capturing more paraphrases. The ensemble approach combines these complementary signals - precise explicit matches and broader referential coverage - alongside linguistic features to make more robust predictions about entity salience.  Although the final micro F1 score of 0.98 seems very high, we note that this is in part due to the frequency of negative judgments (most document mentions do not appear in summaries), and the positive class F1 is lower, at 0.9. We also note that the task is easier than might be expected, since our guidelines explicitly ask summary writers to adhere to things mentioned in the documents, and to keep phrasing similar to the source material.

\section{Predicting salient entities}
The purpose of the dataset created in the previous sections is to train and evaluate systems on the task of gradient entity salience prediction. 
To evaluate contemporary models on this task, we prompt\footnote{Settings and prompt in Appendix \ref{sec:appendix-eval}} LLMs to extract (i) salient entities from a document and (ii) a salience score from 1 to 5 for each predicted salient entity (with entities absent from the prediction corresponding to a score of 0). 
\subsection{Models}
\paragraph{Zero-shot/Few-shot LLMs}
We used GPT-4o \cite{openai2024gpt4} in both zero-shot and few-shot settings to identify salient entities within a document and assign a salience score between 1 and 5 to each predicted entity. For the few-shot setting, we provide 3 randomly selected documents from the dev set, along with their salience scores, as in-context examples to guide the model's predictions. Additionally, we evaluated other instruction-based models, including \texttt{Llama-3.2-3B-Instruct} \cite{meta2024llama3} and \texttt{Mistral-7B-Instruct-v0.3} \cite{jiang2023mistral7b}, which are tuned for tasks like information retrieval and text summarization. These models are prompted to extract salient entities, such as people and organizations, and assign salience scores from 1 to 5. We chose instruction-based models over base models due to their fine-tuning on retrieval tasks, which aligns with our goal of retrieving and ranking salient entities.

\paragraph{Stanza \& Ensemble Approach}
We evaluated two approaches: the Stanza model and our proposed ensemble method. Following our definition of salient entities as those present in both documents and summaries, we extracted binary salience labels from each summary and aggregated them across all five model-generated summaries\footnote{For fairness, we used 5 model-generated summaries for evaluation (Section~\ref{sec:eval}), using the same models and methods from Section \ref{sec: sum-crowdsourcing-generation}, despite having human-written ones available. This allows us to evaluate how well our approaches handle end-to-end salience prediction from documents alone, ensuring a fair comparison with LLMs using the same inputs.} to derive a 1-5 salience score per entity. 

\paragraph{Position Baseline}
We built our baseline model on the simple and naive assumption that entities mentioned earlier in a document tend to be more salient, as important information is often introduced early \cite{dojchinovski2016crowdsourced, liu-etal-2018-automatic}. Entities are assigned salience scores based on their first appearance within the document. The document is divided into sections, where entities in the first 10\% of sentences are scored 5, with scores decreasing by 1 for each subsequent 10-20\% segment, down to 0 for the last 20\%. This method provides a simple, interpretable baseline that reflects the salience of entities based on their position in the document.

\subsection{Evaluation Metrics}

In our evaluation experiment, we evaluate the correlation between predicted and human-aggregated salience scores using two metrics: Spearman's $\rho$ and Root Mean Square Error (RMSE), which capture different aspects of prediction quality. Spearman's $\rho$, ranging from -1 to +1, measures the strength and direction of the monotonic relationship between the rankings of predicted and human scores. For example, if two entities are ranked similarly by both the model and the annotators (even if the raw scores differ), we will observe a high Spearman correlation. In contrast, RMSE measures the absolute difference between predicted and annotated scores, providing insight into how far off the predictions are on average, regardless of the ranking; if a model consistently predicts scores that are close to the annotated values, RMSE will be low. \par

To evaluate the effectiveness of the models in identifying entities with high salience, we compute precision, recall, and F1 scores for the top1 (entities with a score of 5 out of 5) and top3 (entities with a score of 3, 4, 5 out of 5) entity ranks to capture the model's ability in detecting the most salient entities. We expect a positive Spearman $\rho$, indicating a correlation in ranking performance. For precision, recall, and F1 scores, we expect the highest scores for top1 entities, with the scores decreasing as we move to top3. This is because we expect models to be more accurate at identifying highly salient entities (score 5), and to degrade as the salience level becomes less distinctive.

\section{Evaluation}
\label{sec:eval}
In this section, we evaluate model performance on the graded entity salience prediction task end to end, measuring the correlation between predicted and human-aggregated salience scores using only the document as input (Section \ref{sec: llm-eval}). In Section \ref{sec: error-analysis}, we focus on error analysis using GPT4o predictions; analyses and predictions of other models are included in Appendix~\ref{sec:appendix-add-analysis}.
\subsection{Model Performance}
\label{sec: llm-eval}

The results in Table~\ref{tb:sp_RMSE} show that the Ensemble approach achieves the highest Spearman’s $\rho$ (0.54), indicating the strongest alignment with human salience rankings, while Stanza achieves the lowest RMSE (1.03). Both methods outperform GPT-4o in zero-shot ($\rho$: 0.229, RMSE: 1.143) and three-shot settings (0.254, 1.111), suggesting that despite their generalization capabilities, LLMs may struggle with fine-grained salience prediction tasks compared to task-specific methods. Additionally, Wilcoxon signed-rank tests confirm that the improvements in rank correlation over GPT-4o (3-shot) are statistically significant for both Stanza and Ensemble (p < 0.001), supporting the robustness of our approach. While these results are promising, with Ensemble showing substantial alignment with rankings based on human-generated summaries and alignments, and Stanza's predictions deviating by approximately one point on the 0 to 5 scale, there remains room for improvement on this challenging task.

\label{Sec:eval}
\begin{table*}[h!]
    \centering
    \scalebox{0.8}{
    \begin{tabular}{lccc}
        \toprule
        & \textbf{Spearman $\rho$ ( 95\% CI)} & \textbf{RMSE ( 95\% CI)} & \textbf{Wilcoxon Test (Spearman $\rho$ vs GPT4o 3-shot)} \\
        \midrule
        Position Baseline & 0.153 (0.103, 0.208) & 2.554 (2.471, 2.636) & Spearman: {*} (\textit{p}<0.05)	\\
        GPT4o 3 shot & 0.254 (0.208, 0.300) & 1.111 (1.044, 1.182) & -	\\
        GPT4o zero shot & 0.229 (0.179, 0.280) & 1.143 (1.075, 1.213) & ns\\
        Llama-3.2-3B-Instruct & 0.223 (0.167, 0.281) & 1.296 (1.211, 1.397) & ns	\\
        Mistral-7B-Instruct-v0.3  & 0.254 (0.202, 0.307) & 1.206 (1.124, 1.286) & ns	\\
        \textbf{Stanza (our approach)} & 0.384 (0.324, 0.437) & \textbf{1.031} (0.932, 1.131) & Spearman: {***} (\textit{p}<0.001) \\
        \textbf{Ensemble (our approach)} & \textbf{0.540} (0.486, 0.589) & 1.067 (0.933, 1.208) & Spearman: {***} (\textit{p}<0.001)\\
        \bottomrule
    \end{tabular}}
    \caption{Performance of all models on the graded entity salience prediction task (test set). We report Spearman's $\rho$ and RMSE, each with 95\% confidence intervals. Our two approaches—Stanza and Ensemble—significantly outperform GPT-4o (3-shot) in rank correlation (Spearman's $\rho$), as confirmed by a Wilcoxon signed-rank test (\textit{p} < 0.001). The Ensemble model achieves the strongest overall performance. Best scores per column are shown in \textbf{bold}. ``ns'' indicates a non-significant difference at \textit{p} $\geq$ 0.05.}
    \label{tb:sp_RMSE}
\end{table*}

\begin{table}[ht]
\centering
\scalebox{0.8}{
\begin{tabular}{lccc}
\toprule
Model & \textbf{P@top1} & \textbf{R@top1} & \textbf{F@top1} \\
\midrule
GPT4o 3 shot & \textbf{0.548} & 0.321 & 0.405 \\
GPT4o & 0.541 & 0.377 & \textbf{0.444} \\
Mistral 7b & 0.415 & 0.321 & 0.362 \\
Llama 3-2 & 0.302 & 0.359 & 0.328 \\
Stanza & 0.239 & 0.491 & 0.321 \\
Ensemble & 0.242 & \textbf{0.755} & 0.367 \\
Position Baseline & 0.031 & 0.208 & 0.054 \\\midrule
Avg & 0.331 & 0.405 & 0.326 \\
\midrule
Model & \textbf{P@top3} & \textbf{R@top3} & \textbf{F@top3} \\
\midrule
GPT4o 3 shot & 0.278 & 0.513 & 0.361 \\
GPT4o & 0.255 & 0.481 & 0.333 \\
Mistral 7b & 0.205 & 0.455 & 0.283 \\
Llama 3-2 & 0.183 & 0.442 & 0.259 \\
Stanza & 0.424 & 0.474 & 0.448 \\
Ensemble & \textbf{0.463} & \textbf{0.610} & \textbf{0.527} \\
Position Baseline & 0.031& 0.305 & 0.057 \\\midrule
Avg & 0.263 & 0.469 & 0.324 \\
\bottomrule
\end{tabular}}
\caption{Precision, recall, and F1 scores for all models on the test set. @Top1 means only the entities with a score of 5 are considered; @Top3 means the entities with a score of 3 or 4 or 5 are considered. The highest scores in each scenario are in \textbf{bold}.}
\label{tb:graded_prf}
\end{table}

Table~\ref{tb:graded_prf} evaluates the model's performance in identifying salient entities under two evaluation settings: Top1 (only entities with a salience score of 5) and Top3 (entities with a salience score of 3, 4, or 5). Importantly, these evaluations are restricted to salient entities only and do not include non-salient entities (salience score = 0) as negatives. As such, the reported scores reflect the model's ability to prioritize the most salient entities without being affected by the large number of non-salient mentions in the dataset.

Overall, the Top1 setting emphasizes strict precision on highly salient entities, whereas Top3 favors a more recall-oriented evaluation. Precision is higher under the Top1 setting, where only the most salient entities are considered, while recall increases under the more inclusive Top3 setting. This trade-off arises because the Top3 evaluation rewards models for capturing additional, moderately salient entities (scores 3 or 4), but at the cost of introducing more false positives, thereby reducing precision.

Our Stanza and Ensemble approaches excel in recall for Top1 salient entities, with the Ensemble achieving the highest recall (0.755). This is because our methods are more lenient, aiming to capture as many salient entities as possible, which increases recall but also introduces false positives, thereby lowering precision. In contrast, LLMs like GPT4o are more conservative in their predictions, focusing on a narrower set of entities that are clearly salient. This conservatism results in higher precision (e.g., 0.548 for GPT4o few-shot) but inherently limits their recall, as they may overlook less obvious yet salient entities.

For the Top3 scenario, our Ensemble approach outperforms all models, achieving the highest recall (0.61) and F1 score (0.527), effectively balancing precision and recall to capture moderately salient entities. The Stanza method also performs well, achieving the second-highest precision (0.424) among all models and a recall (0.474) that outperforms several LLMs. In contrast, LLMs like GPT4o underperform, with both precision and recall falling short of the Ensemble, highlighting the advantage of our task-specific strategies in identifying a wider range of salient entities.

Although our Ensemble approach achieves the highest performance in the Top3 scenario, with an F1 score of 0.52 and a spearman correlation of 0.54, these results highlight the fundamental challenge of consistently identifying and ranking entities by their salience in text.

\subsection{Error Analysis}
\label{sec: error-analysis}
\begin{figure}[h!]
\centering
\includegraphics [width=0.4\textwidth] {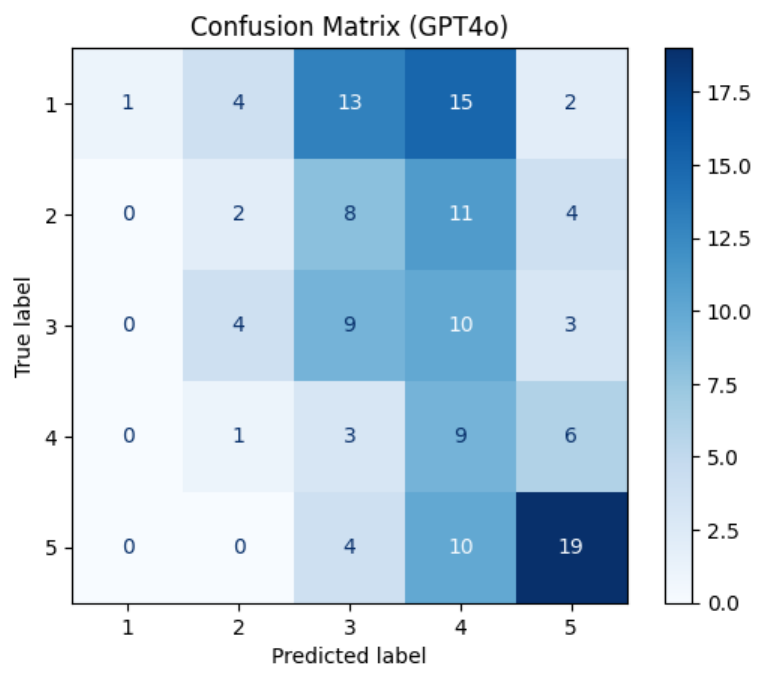}
\caption{Confusion matrix for GPT4o}
\label{fig:cm_GPT4o}
\vspace{-12pt}
\end{figure}

The confusion matrix in Figure~\ref{fig:cm_GPT4o} shows that GPT4o is more confident in predicting highly salient entities but struggles to differentiate less salient ones, often overpredicting moderate salience scores (3 and 4) for entities that are actually less salient (1 and 2). This pattern is observed in other models as well (Figure~\ref{fig:cm_all_models}). The model performs best with highly salient entities (score 5) but tends to misclassify low-salience entities (score 1) into higher salience categories like 3 or 4, likely because moderate scores are a safer prediction for uncertain cases, reducing extreme errors but introducing more moderate misclassifications. We speculate that models' natural language understanding does not calibrate them well to the scale implied by the data.\par

\begin{figure}[h!]
\centering
\includegraphics[width=0.5\textwidth] {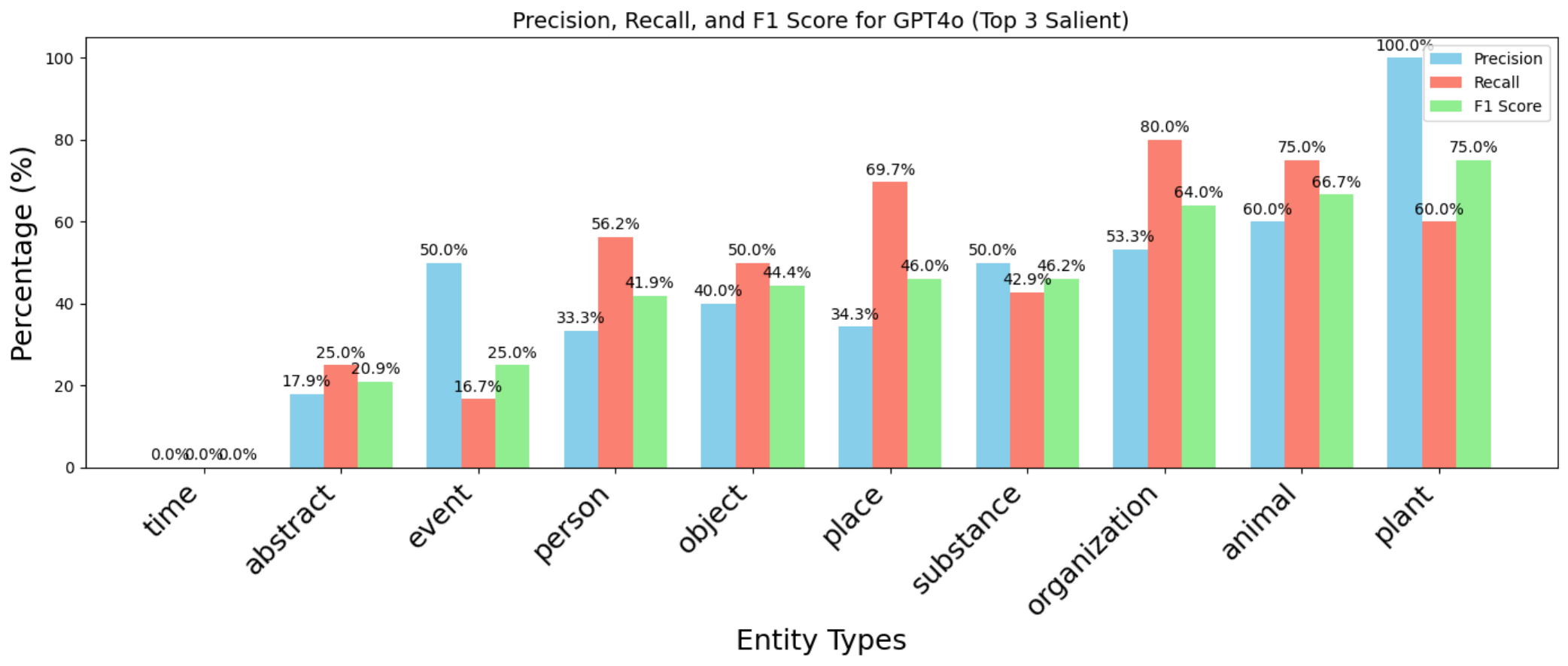}
\caption{Performance of GPT4o across entity types in identifying the top 3 salient entities. Precision (the blue bar) measures the percentage of predicted salient entities for a given type that are correct, recall (the red bar) measures the percentage of actual salient entities that the model successfully predicts.}
\label{fig:EntTypeErrorRateTop3}
\end{figure}

The results in Figure~\ref{fig:EntTypeErrorRateTop3} shows GPT4o's varying performance across entity types when predicting the Top 3 salient entities. The model performs well on concrete and named entities like \textsc{animal}, \textsc{plant} and \textsc{organization}, which exhibit higher precision, recall, and F1 scores.
Conversely, the model struggles with conceptual or contextual entities such as \textsc{abstract}, \textsc{event}, and \textsc{time}, reflecting a need for deeper semantic understanding to predict salience effectively. For entities like \textsc{person} and \textsc{place}, the much higher recall compared to precision suggests overpredictions for these frequently occurring and easily identifiable entity types. This performance disparity\footnote{Similar patterns can be seen in other models as well (Figure~\ref{fig:prf_all_models}).} highlights the model's strengths with concrete entity types but its limitations in processing context-dependent or abstract entities. \par
\begin{figure}[h!]
\centering
\includegraphics[width=0.5\textwidth] {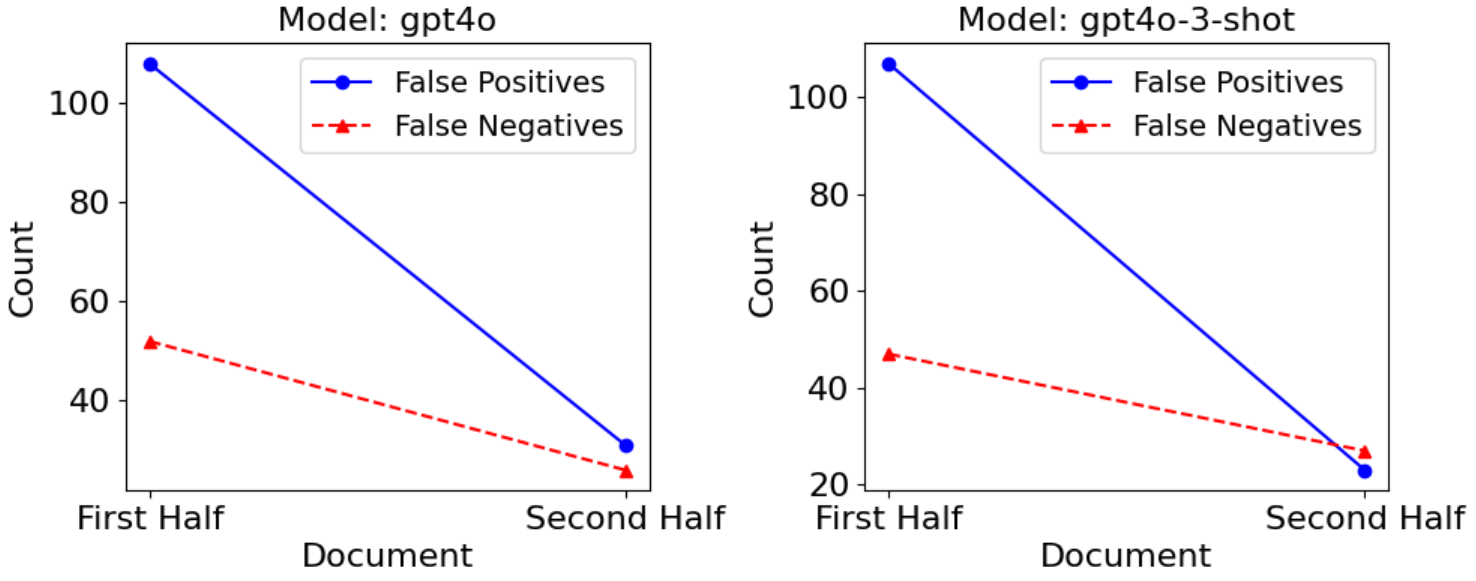}
\caption{Interaction plot showing the counts of false positives (FP) and false negatives (FN) for GPT4o models across the first and second halves of documents.}
\label{fig:EntPosFpFn}
\end{figure}

The results in Figure~\ref{fig:EntPosFpFn} show that for GPT4o models, false positive (FP) and false negative (FN) counts decrease from the first to the second half of the document, with FP counts declining more steeply. This indicates that the model overpredicts salience for entities mentioned early in the document, resulting in a higher FP rate in the first half. The steep FP decline suggests models are more conservative in the second half, likely due to fewer entities being introduced. The gentler FN decline reflects the smaller number of salient entities in the second half, giving models fewer opportunities to miss salient predictions.

\section{Conclusion}
\label{sec:5}
In this paper, we introduced a novel approach to graded salient entity extraction, combining the strengths of human judgments and summary-based operationalization. By aligning multiple summaries for each document to their texts, we created a dataset with graded salience scores that balances gradient scoring with consistency and explainability. While predicting entity salience remains challenging even for powerful LLMs in zero-shot and few-shot settings, our novel SEE approach demonstrates promising results by outperforming these models. Our analysis shows that models perform well on concrete entities but poorly on abstract ones (abstract notions, time), with bias toward common entities and those appearing early in text. 

The data set and approach presented here establish a foundation for improving salience modeling in summarization and information retrieval.  We are also planning to add the same graded annotations described in this paper to upcoming editions of the GUM corpus currently covering 16 genres (with the addition of court transcripts, essays, letters and podcasts), as well as the out-of-domain challenge test set GENTLE (the Genre Tests for Linguistic Evaluation corpus, \citealt{aoyama-etal-2023-gentle}), with eight additional genres.
Future work using this data should continue to address model biases and promote stability across genres.


\section*{Limitations}
Our work has several limitations. First, the graded summary-based approach relies on multiple high-quality summaries per document (either human- or model-generated), which introduces scalability challenges—particularly in low-resource settings. Collecting and validating multiple summaries is resource-intensive and naturally constrains dataset size. While this setup is appropriate for evaluation and for establishing the gradient-based salience framework, we anticipate that human-written summaries may not be required when deploying the method in practice. As shown in Table~\ref{tb:sp_RMSE} and \ref{tb:graded_prf}, our methods perform competitively using only model-generated summaries. Moreover, recent advances in APIs have made it easier to generate multiple summaries at scale, reducing the overhead associated with querying different LLMs.

Second, we acknowledge the potential for pretraining contamination in LLMs, as our evaluation documents come from public sources. Specifically, since our dataset uses the open-source GUM corpus, some documents might have been part of models like GPT-4's pretraining data. However, we believe this concern primarily affects the evaluation of summarization quality, not the alignment-based salience prediction task. The summaries used for alignment are independently generated (by humans or models), and the alignment process involves matching \textit{novel} summary content to entities in the source text, which makes it unlikely that the exact summary-document pairs were observed during pretraining. Furthermore, as shown in Figure~\ref{fig:self-bleu}, the low Self-BLEU scores across genres indicate that the summaries in our dataset are lexically diverse and not overly repetitive, supporting their effectiveness as a robust input for SEE even in the presence of potential pretraining overlap.

Third, our dataset is currently limited to English, the highest-resource language in NLP. The performance of pretrained models on both summarization and entity salience tasks would likely be substantially lower for other languages. While we
believe the fundamental approach of using multiple summaries to grade entity salience should generalize across languages, this remains to be empirically verified, particularly for languages with different discourse structures or conventions around
entity reference.

Finally, although the dataset spans 12 genres, domain-specific patterns of entity salience may still be underrepresented. The semi-automatic alignment process, while scalable and robust, continues to require human validation to ensure accuracy. This may limit applicability in genres or domains where reference patterns or discourse organization differ substantially from those represented in our corpus.

\section*{Acknowledgments}
The summary crowdsourcing study was supported by a GSAS-GradGov Research Project Award (GRPA), which funds graduate student research and professional development at Georgetown University. We are grateful to the following participants for their valuable contributions and thoughtful feedback during the summary crowdsourcing process (listed alphabetically by last name): Caroline Coggan, Jessica	Cusi, Dan DeGenaro, Caroline Gish, Aniya Harris, Abby Killam, Lauren Levine, Cindy Li, Robbie Li, Cindy Luo, Todd McKay, Sophie Migacz, Anna Prince, Emma Rafkin, Eliza Rice, Wesley Scivetti, Devika Tiwari, Shira Wein.


\bibliography{anthology, custom}
\bibliographystyle{acl_natbib}

\appendix


\section{API costs}
\subsection{Summary Generation}
For summary generation, we produced four one-line summaries for each of the 165 documents in the training set, totaling 660 requests. Using GPT-4o, priced at \$2.50 per 1M input tokens and \$10.00 per 1M output tokens\footnote{\url{https://openai.com/api/pricing/}}, the cost was \$0.90. For Claude 3.5 Sonnet, priced at \$3 per 1M input tokens and \$15 per 1M output tokens\footnote{\url{https://www.anthropic.com/pricing\#anthropic-api}}, the cost was \$3.34. In total, the cost for summary generation was \$4.24. Generating summaries with Llama 3.2 and Qwen models using the Huggingface \texttt{transformers}\footnote{See \url{https://huggingface.co/meta-llama/Llama-3.2-3B-Instruct} for the Llama model and \url{https://huggingface.co/Qwen/Qwen2.5-1.5B-Instruct} for the Qwen model used in this paper.} library did not incur any additional costs, as these models are open-source and locally hosted.
\subsection{Automatic Alignment}
The alignment task involved matching entities across document-summary pairs, requiring 6,660 requests processed using GPT-4o, with a total cost of \$40.14. 

\section{Summary Generation \& Crowdsourcing Details} \label{sec:appendix-sc-details}

\subsection{Summary Generation}
\label{sec:appendix-sg}
\begin{figure}[htb]
\centering
\includegraphics [width=0.48\textwidth] {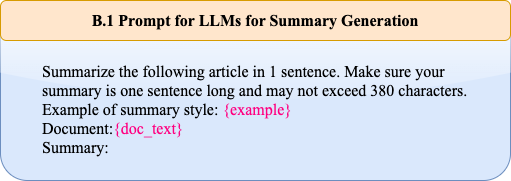}
\caption{Prompt for LLMs for Summary Generation.}
\label{fig:SumGenPrompt}
\end{figure}

Figure~\ref{fig:SumGenPrompt} shows the prompt to generate a one-sentence summary with all LLMs (\texttt{GPT4o}, \texttt{Claude 3.5 Sonnet}, \texttt{Llama 3.2 3B}, \texttt{Qwen 2.5 7B}). A max token of 120 is used to make sure the generated summary is not too long. All the other hyperparameters are in their default settings.
\subsection{Summary Crowdsourcing}
\label{sec:appendix-sc}
\begin{figure}[htb]
\centering
\includegraphics [width=0.48\textwidth] {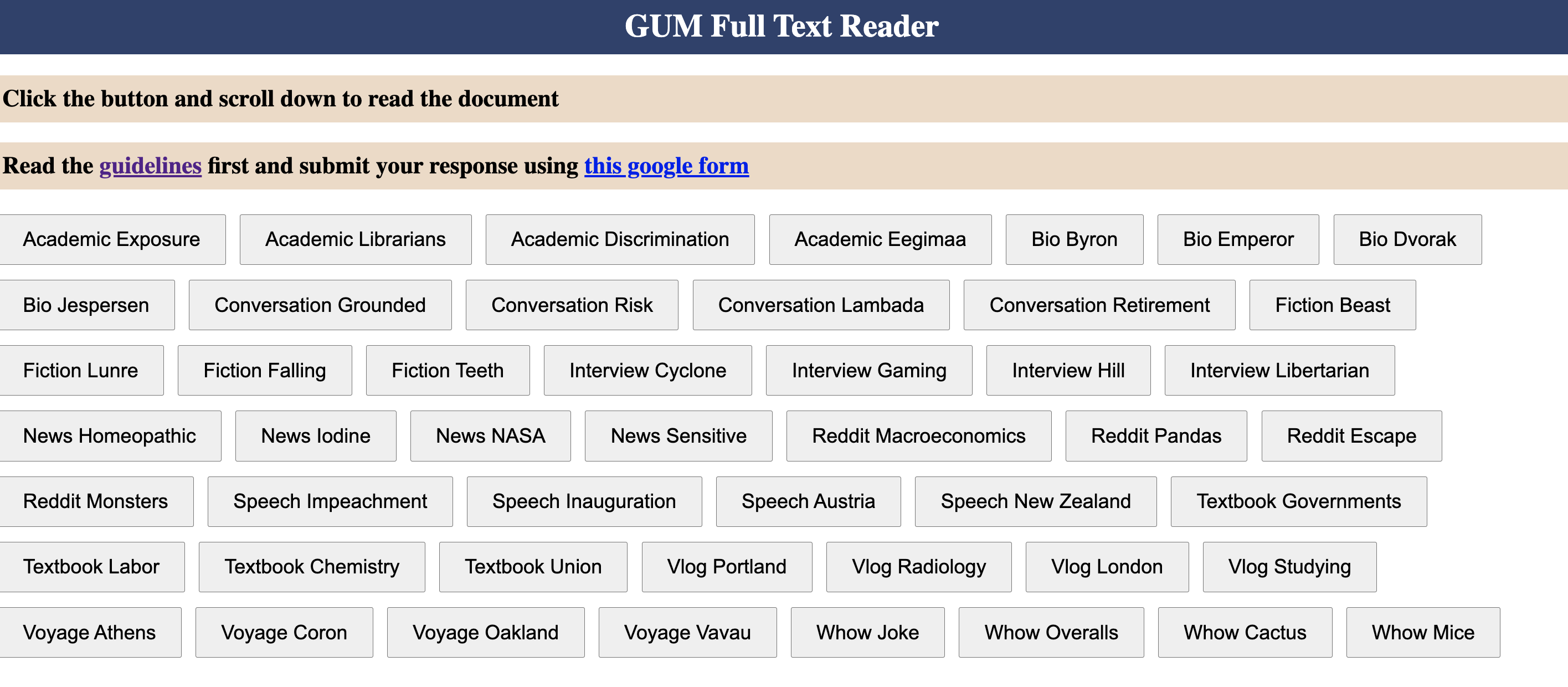}
\caption{GUM Full Text Reader: The interface for viewing GUM articles.}
\label{fig:AnnoInterface}
\end{figure}

\begin{figure}[htb]
\centering
\includegraphics [width=0.48\textwidth] {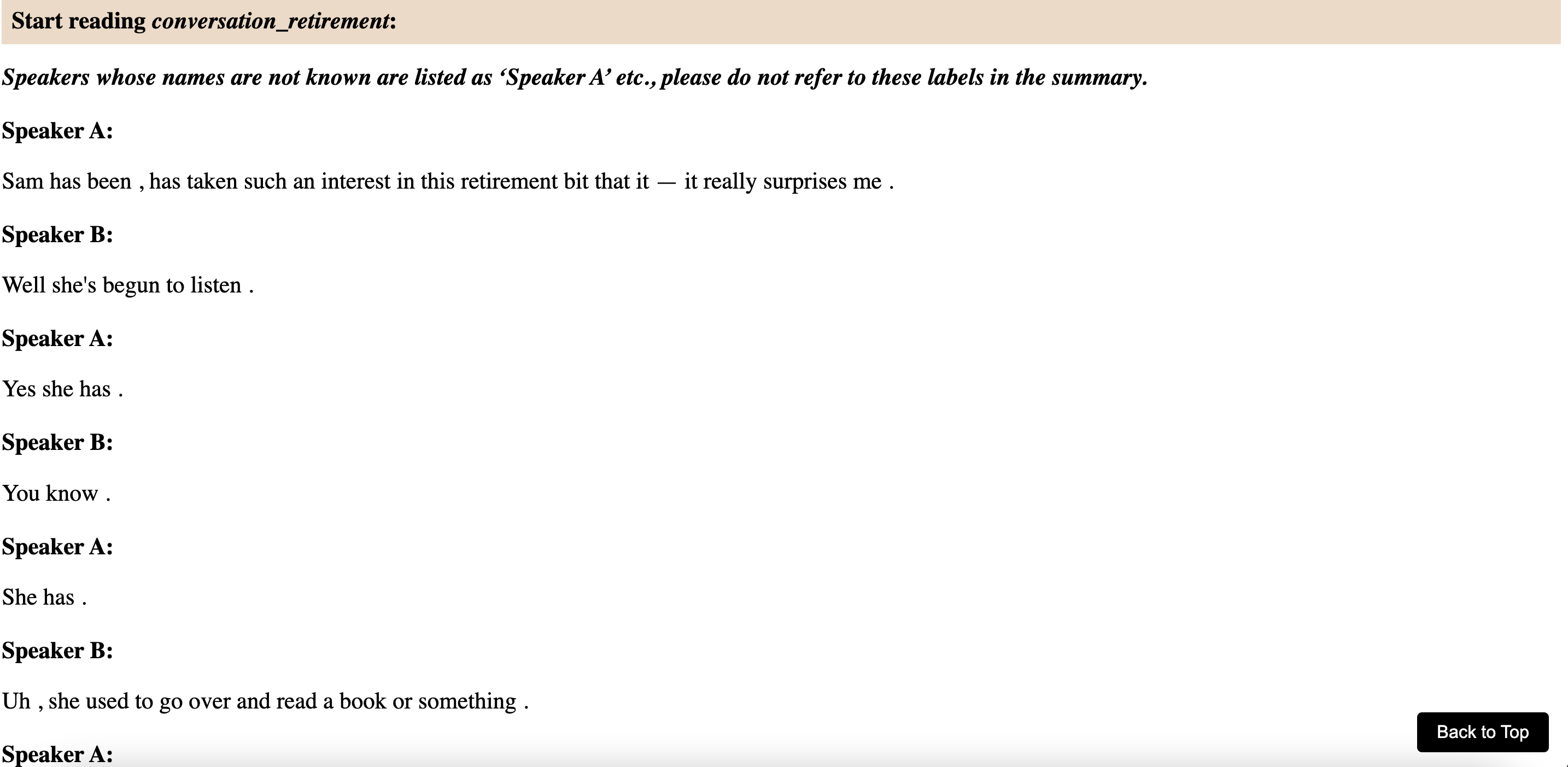}
\caption{An example of GUM Full Text Reader.}
\label{fig:gumView}
\end{figure}

\begin{figure}[htb]
\centering
\includegraphics [width=0.48\textwidth] {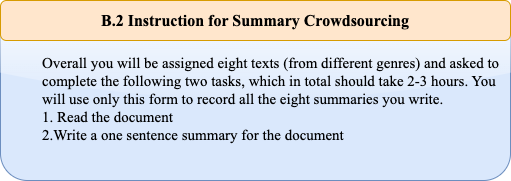}
\caption{Instruction for summary crowdsourcing.}
\label{fig:Inst-overall}
\end{figure}

\begin{figure}[htb]
\centering
\includegraphics [width=0.48\textwidth] {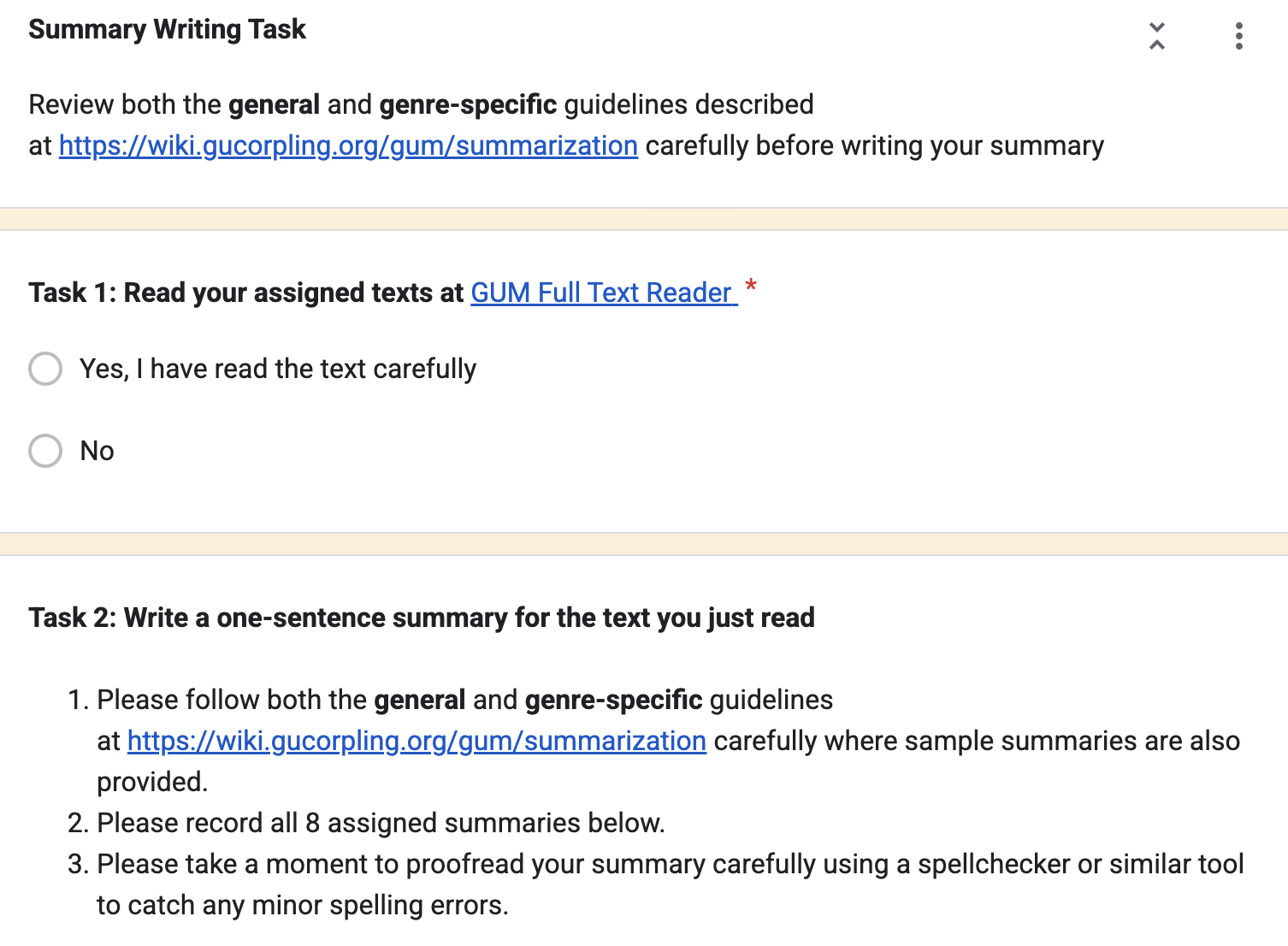}
\caption{A screenshot of instructions given to the annotators.}
\label{fig:Inst-2}
\end{figure}

We recruited 24 graduate students at Georgetown University who are native speakers of English to write summaries for the test and dev set of \corpname{}. Each writer, paid \$22/hr (based on the pay rate
of the 2023 / 2024 academic year for graduate students at Georgetown University), wrote 8 summaries, resulting in 192 human-written summaries in total. Each student received a Google Form to write summaries for their assigned texts, which can be viewed from an interface titled \texttt{GUM Full Text Reader} (Figure~\ref{fig:AnnoInterface} and \ref{fig:gumView}). Figure~\ref{fig:AnnoInterface} and \ref{fig:gumView} show the interface for viewing articles in GUM. Figure~\ref{fig:Inst-overall} and \ref{fig:Inst-2} show the instructions given to the writers.

\section{Automatic Evaluations of Summary Quality}
\label{sec:appendix-auto-sumeval}
\begin{figure*}[tb]
\centering
\includegraphics [width=\textwidth] {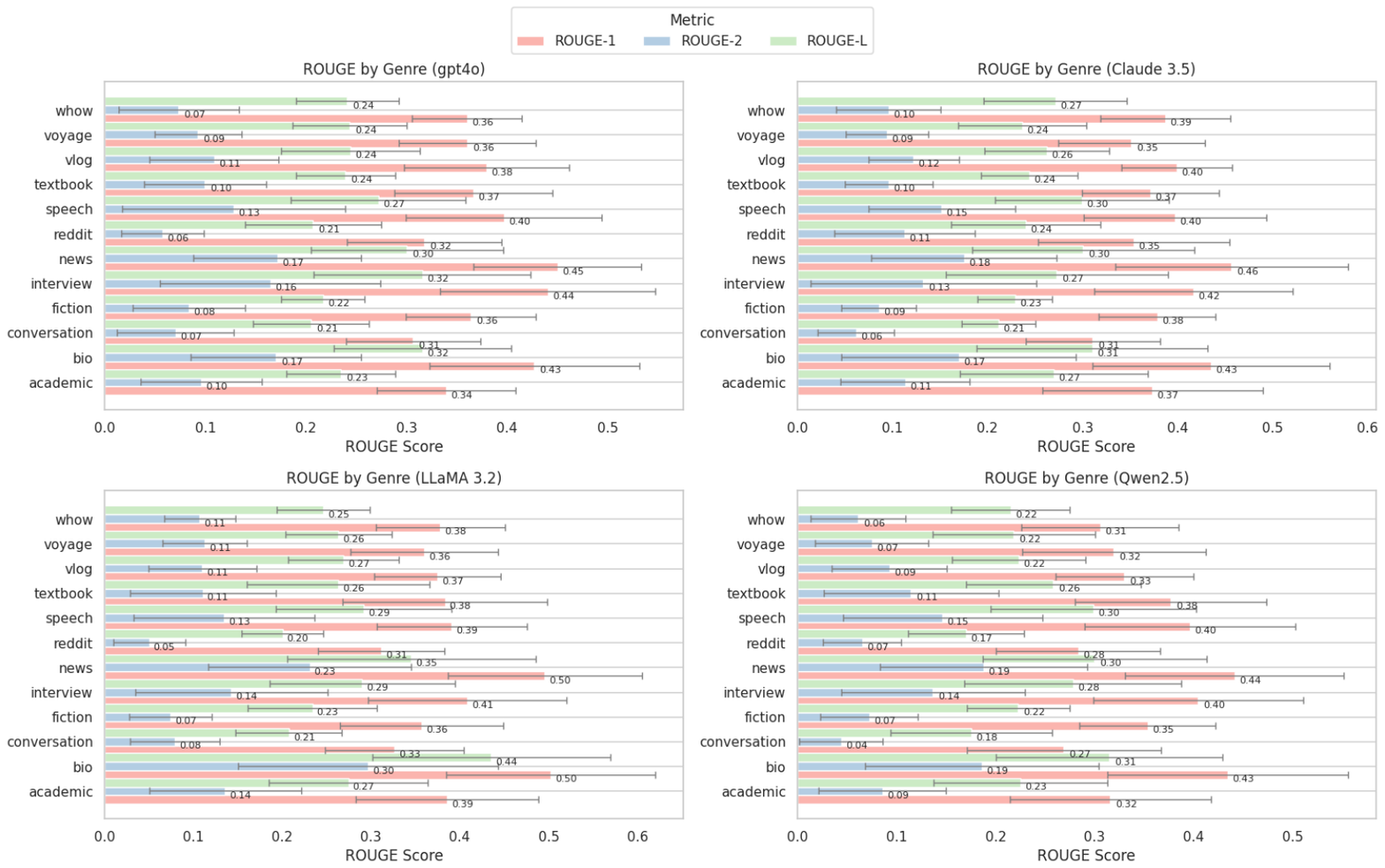}
\caption{ROUGE-1, ROUGE-2, and ROUGE-L scores (with standard deviation) for summaries generated by four models (\texttt{GPT-4o}, \texttt{Claude 3.5}, \texttt{LLaMA 3.2}, and \texttt{Qwen2.5}), evaluated against human-written summaries across genres in the \texttt{training} set. Each subplot shows genre-level performance for one model.}
\label{fig:rouge}
\end{figure*}

To validate the quality of the summaries used for salient entity alignment in Section~\ref{sec: auto-alignment}, we conducted automatic evaluations comparing the human-written and LLM-generated summaries across genres. ROUGE-1, ROUGE-2, and ROUGE-L scores \cite{lin-2004-rouge} were computed using the human summary as reference (Figure~\ref{fig:rouge}). We found that LLM-generated summaries, particularly those from GPT-4o and Claude 3.5, achieve relatively moderate to high ROUGE scores (0.4 to 0.5), especially in genres such as \texttt{news} and \texttt{biography}, suggesting a high degree of lexical overlap with the human reference and thus reliable SEE.

\begin{figure*}[tb]
\centering
\begin{minipage}[t]{0.49\textwidth}
  \centering
  \includegraphics[width=\linewidth]{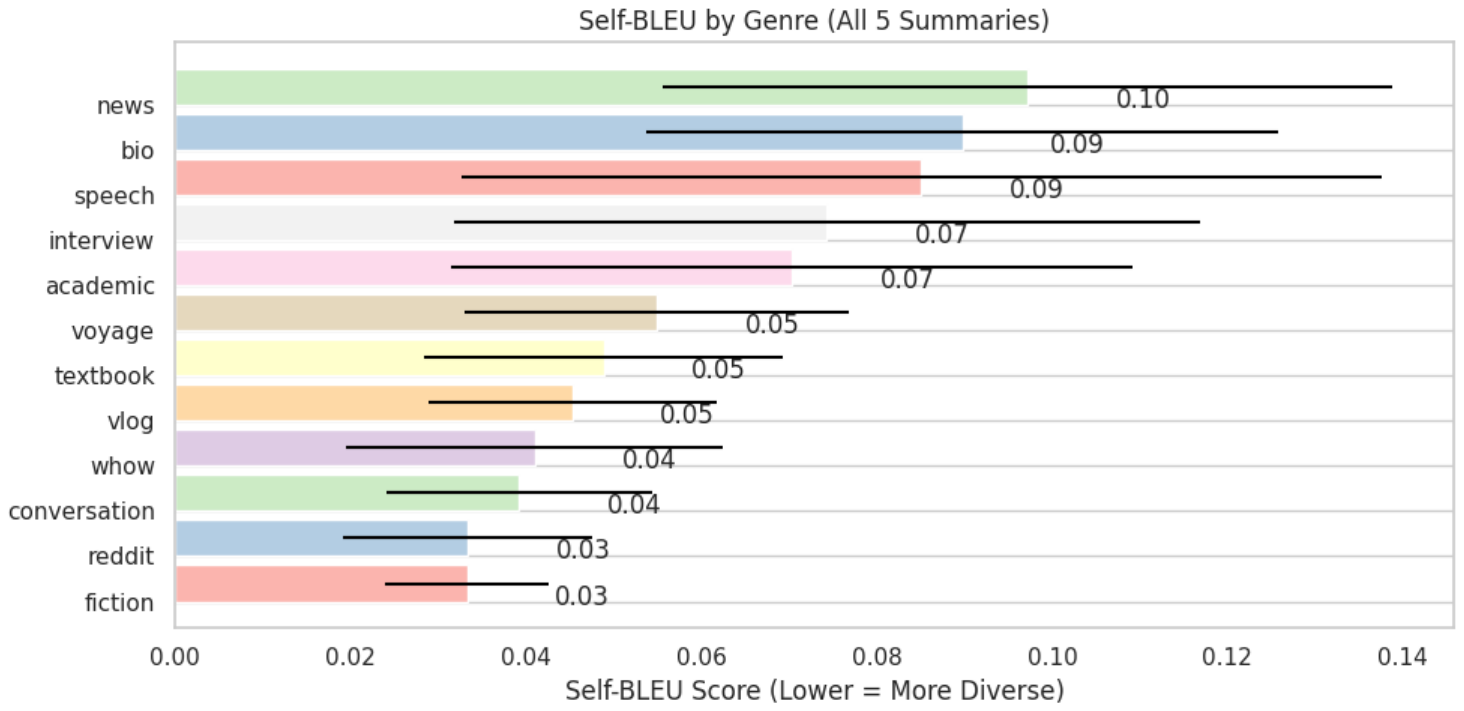}
  \caption{Self-BLEU scores (with standard deviation) for each genre, computed across all five summaries (1 human + 4 model-generated). Lower Self-BLEU values indicate greater lexical diversity among the summaries.}
  \label{fig:self-bleu}
\end{minipage}
\hfill
\begin{minipage}[t]{0.49\textwidth}
  \centering
  \includegraphics[width=\linewidth]{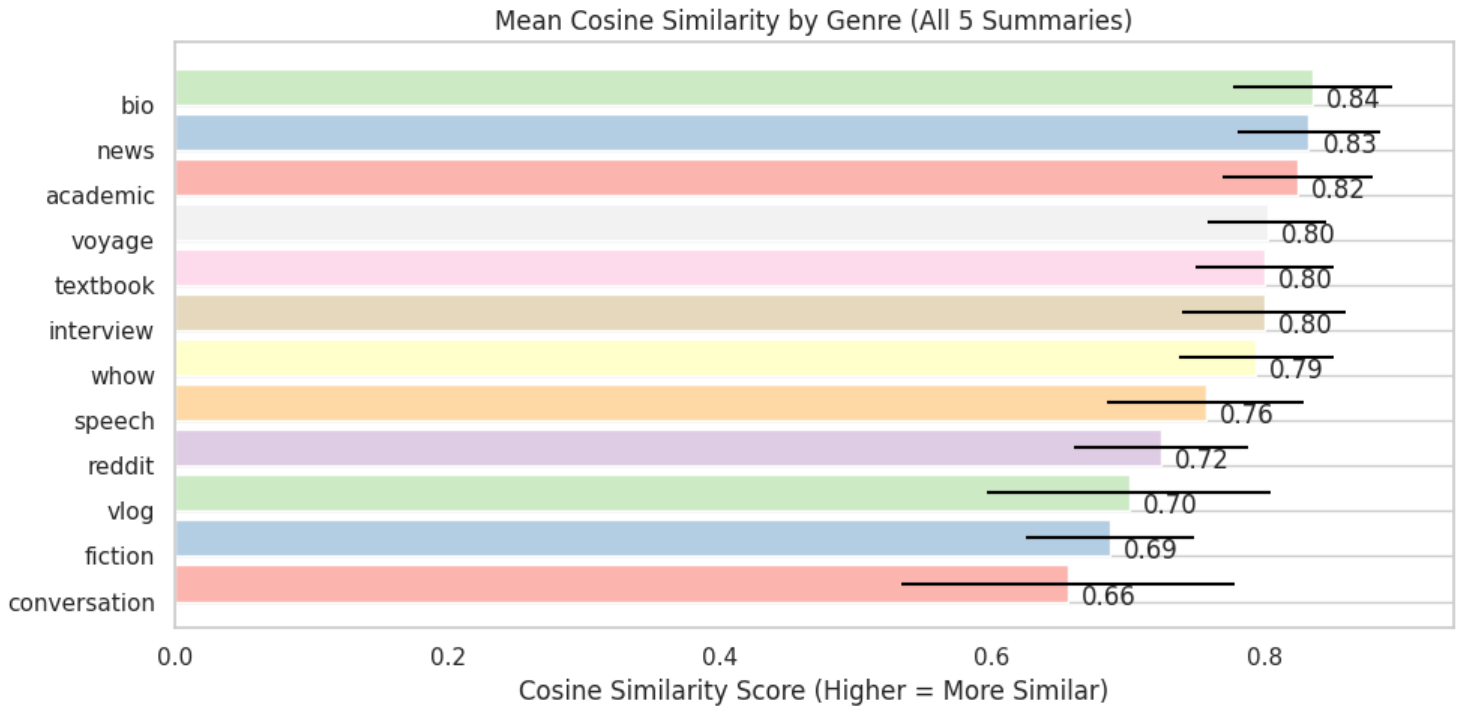}
  \caption{Mean pairwise cosine similarity (with standard deviation) between all five summaries per document, averaged by genre. Cosine similarity captures semantic similarity based on sentence embeddings, with values closer to 1 indicating stronger agreement in meaning.}
  \label{fig:cos-sim}
\end{minipage}

\end{figure*}

We also evaluate the lexical and semantic diversity of the five summaries per document (one human + four model-generated) using Self-BLEU \cite{zhu2018texygen} and mean pairwise cosine similarity, in Figure~\ref{fig:self-bleu} and Figure~\ref{fig:cos-sim} respectively. Self-BLEU measures lexical diversity by averaging BLEU scores across all summary pairs. It ranges from 0 (completely diverse) to 1 (identical), with lower values indicating greater diversity. As indicated in Figure~\ref{fig:self-bleu}, Self-BLEU scores remain low (typically < 0.10) across genres, suggesting that the summaries are lexically diverse and not overly repetitive.

Cosine similarity, computed over sentence embeddings, captures how semantically similar the summaries are. It ranges from –1 to 1, with higher values indicating greater semantic alignment. In Figure~\ref{fig:cos-sim}, we observe high cosine similarity scores (e.g., 0.7-0.8) for most genres, reflecting strong consensus among human and model summaries in what content to include. In contrast, lower similarity scores for genres like \texttt{fiction} and \texttt{conversation} (e.g., 0.66–0.70) suggest a greater variability in what is considered salient in these genres. Together, these results confirm that the summaries used in our pipeline are both high-quality and sufficiently diverse to support robust salient entity extraction.

\section{Summary alignment interface}
\label{sec:appendix-interface}
\begin{figure*}[tb]
\centering
\includegraphics[width=1\textwidth, trim={1.2cm 13.3cm 2cm 1cm}, clip]{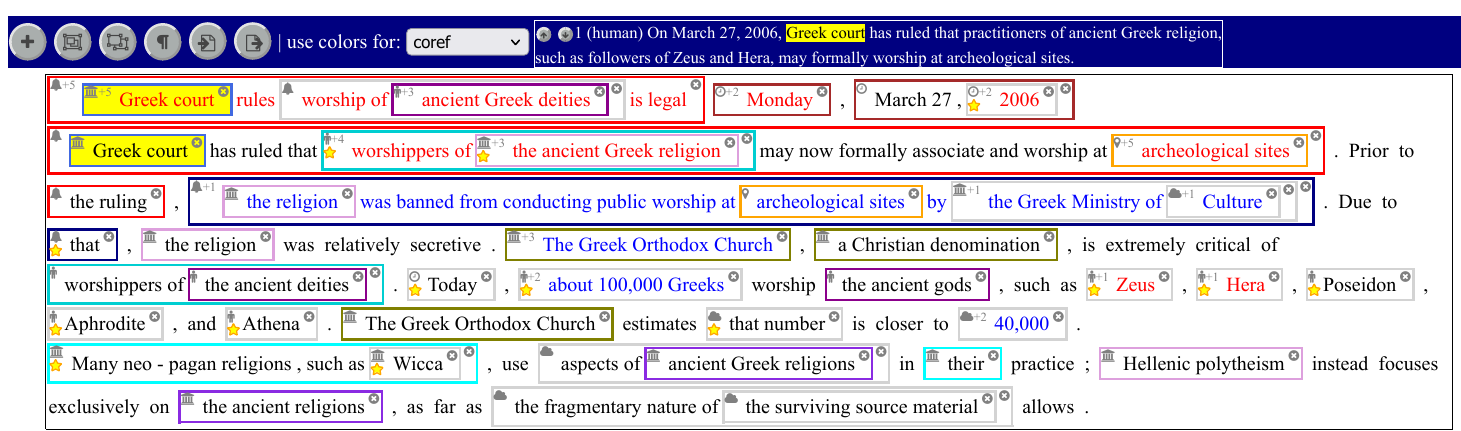}
\caption{GitDOX annotation interface for summary entity alignment.}
\label{fig:spannotator}
\end{figure*}

To inspect, evaluate and correct summary alignment, we used the GitDOX interface \cite{Zhang2017GitDOXAL}, shown in Figure \ref{fig:spannotator}. The summaries are shown at the top of the interface and can be scrolled through. Coreferring mentions receive boxes with identical color borders, and are highlighted in yellow on hover (`Greek Court' in the example). Entities that appear in the currently selected summary have red text at their first mention, for example `Zeus', while entities that are mentioned in a different summary but not the current one have blue text on their first mention. The total number of summaries that an entity is aligned to is shown with a plus and a number indicator, for example `+1' for Zeus, which appears only in summary1, but `+5` for the court, which appears in all five summaries.

\section{LLM Graded Entity Salience Prediction}
\label{sec:appendix-eval}
\begin{figure}[h!]
\centering
\includegraphics [width=0.48\textwidth] {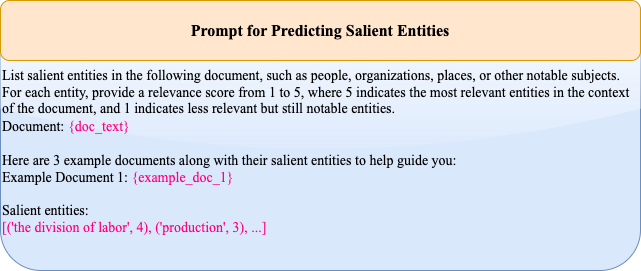}
\caption{Prompt for predicting salient entities (3-shot).}
\label{fig:prompt-eval}
\end{figure}

\begin{figure}[h!]
\centering
\includegraphics [width=0.48\textwidth] {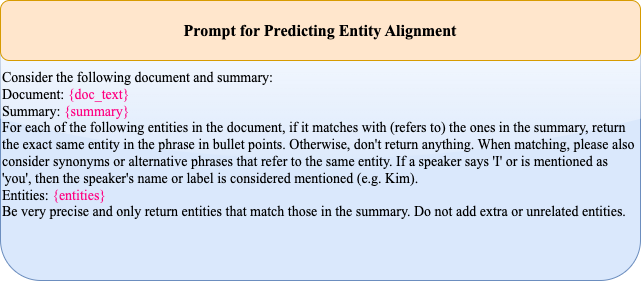}
\caption{Prompt for predicting entity alignment.}
\label{fig:prompt-align}
\end{figure}

We show the prompt for extracting and scoring salient entities in Figure~\ref{fig:prompt-eval}. The prompt for predicting entity alignment is in Figure~\ref{fig:prompt-align}. We set a \texttt{temperature} of 0.2 to encourage deterministic outputs by focusing on the most probable responses. A \texttt{max\_tokens} value of 300 limits verbosity and controls token usage, while a \texttt{top\_p} of 0.7 reduces randomness, ensuring the model prioritizes relevant and reliable predictions. 

\section{Additional Analyses on Graded Entity Salience Prediction}
\label{sec:appendix-add-analysis}
\subsection{LLM Performance across Genres}
\begin{figure*}[htb]
\centering

\begin{minipage}[t]{0.49\textwidth}
  \centering
  \includegraphics[width=\linewidth]{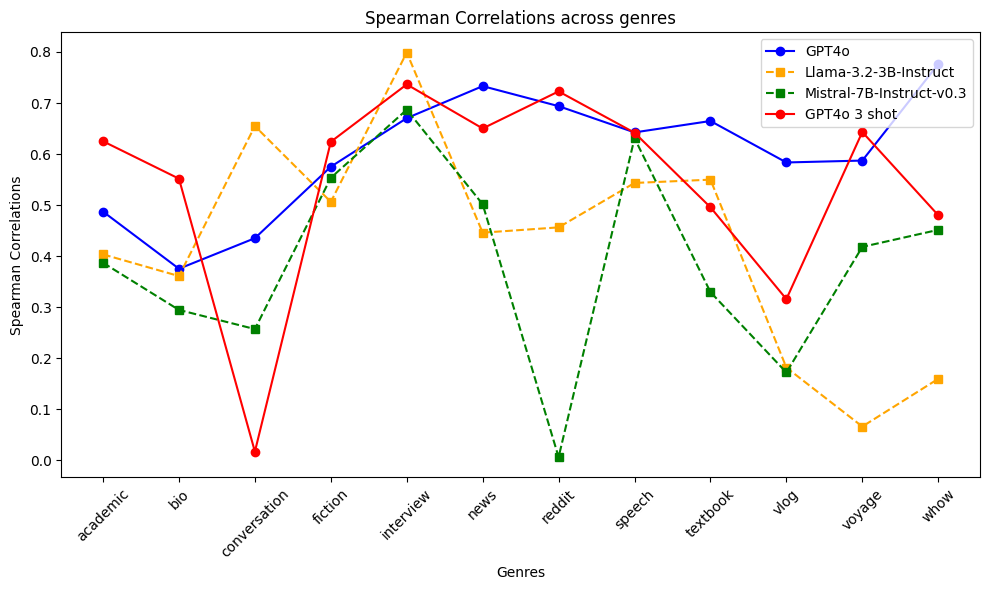}
  \caption{Spearman correlation with LLM scores across 12 genres.}
  \label{fig:SprCorCrossGenre}
\end{minipage}
\hfill
\begin{minipage}[t]{0.49\textwidth}
  \centering
  \includegraphics[width=\linewidth]{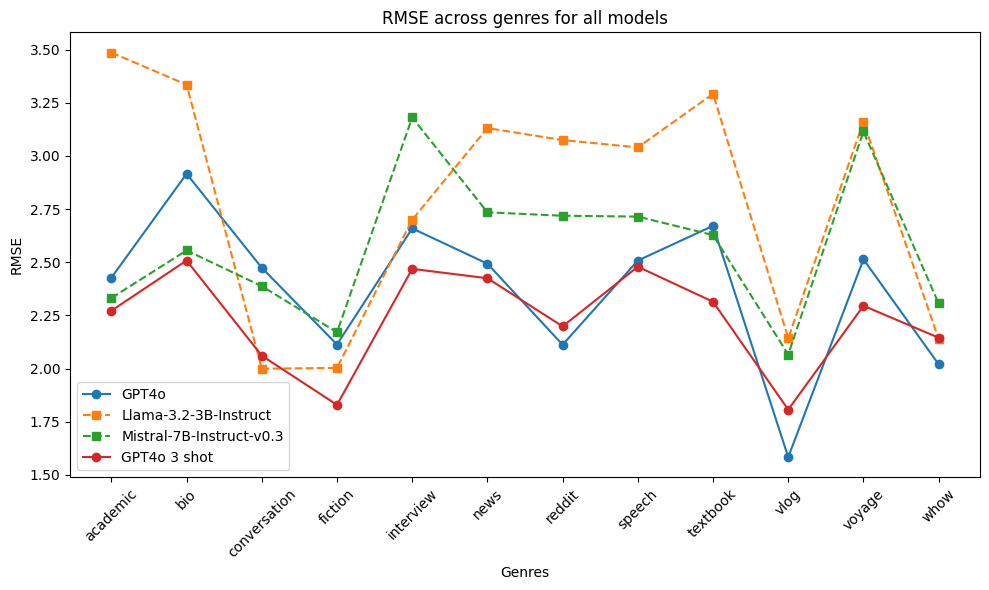}
  \caption{RMSE scores with LLM scores across 12 genres.}
  \label{fig:RMSECrossGenre}
\end{minipage}

\end{figure*}

Figure~\ref{fig:SprCorCrossGenre} shows that all LLMs perform relatively well on structured genres like \texttt{interview}, \texttt{news}, and \texttt{textbook}, which provide clear organizational cues (e.g. section titles and headlines), enabling easier salience scoring. In contrast, performance drops significantly in \texttt{conversation} and \texttt{vlog}, even for powerful models like GPT4o 3 shot, despite being provided with in-context examples. This disparity can be attributed to the unstructured and dynamic nature of these genres: conversations feature fragmented speech, rapid topic shifts, and implicit references, while vlogs often revolve around subjective, informal storytelling. The lack of clear salience signals in these genres makes it challenging for models to align their scores with human judgments.

The RMSE plot in Figure~\ref{fig:RMSECrossGenre} reveals a contrasting trend, where unstructured genres like \texttt{conversation} and \texttt{vlog} perform well, with lower RMSE values compared to structured genres like \texttt{interview} or \texttt{textbook}. This suggests that while models struggle to rank entities accurately in these unstructured genres (as seen in Spearman correlations), they tend to predict scores that are numerically closer to the annotated values. This may occur because unstructured genres often feature a smaller range of salience variation, with fewer highly salient entities and many entities scored similarly. Consequently, even when the ranking is incorrect, the predicted scores remain close to the annotated values, resulting in lower RMSE.

\subsection{Confusion Matrices for all Models}
\begin{figure*}[htb]
    \centering
    \begin{minipage}[b]{0.32\textwidth}
        \centering
        \includegraphics[height=4cm]{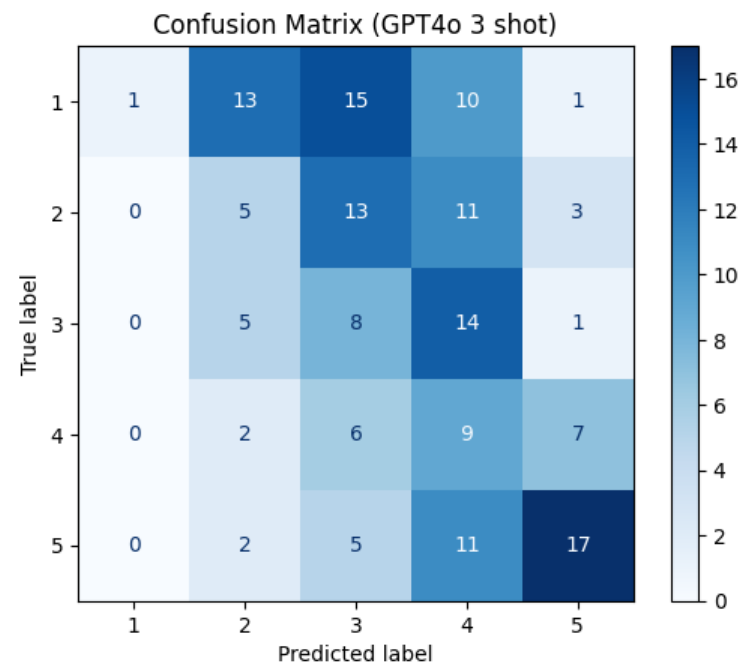}
        \caption*{(a) GPT4o 3-shot}
    \end{minipage}
    \hfill
    \begin{minipage}[b]{0.32\textwidth}
        \centering
        \includegraphics[height=4cm]{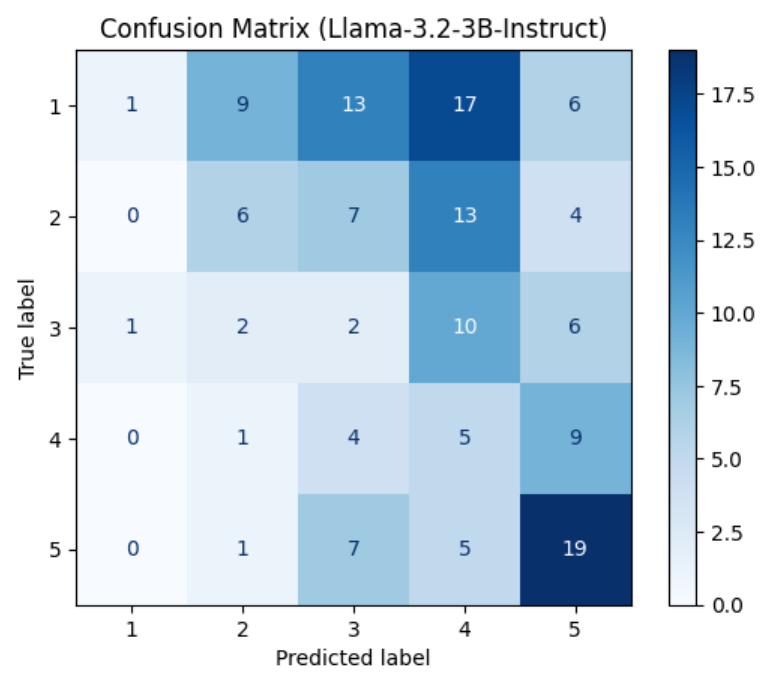}
        \caption*{(b) LLama 3-2}
    \end{minipage}
    \hfill
    \begin{minipage}[b]{0.32\textwidth}
        \centering
        \includegraphics[height=4cm]{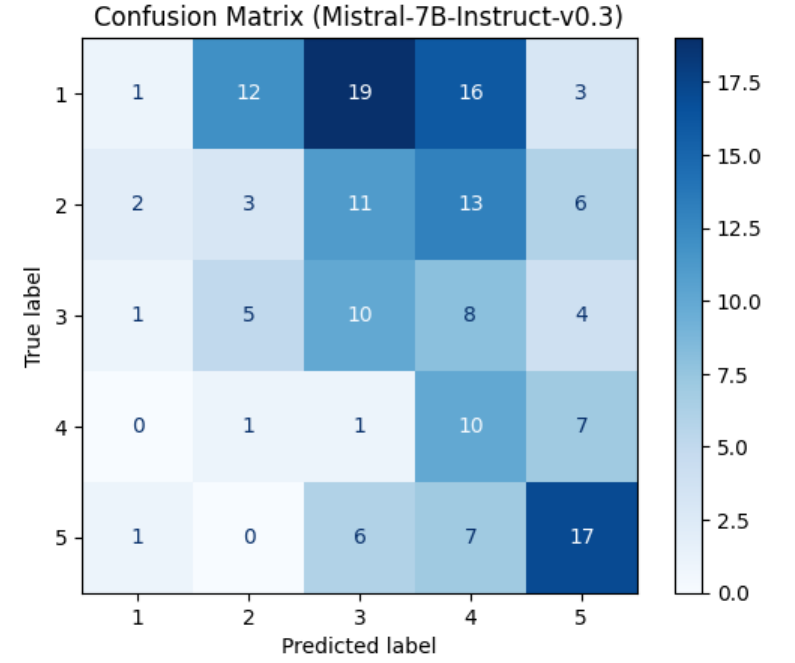}
        \caption*{(c) Mistral 7B}
    \end{minipage}
    \vspace{0.5cm}
    \begin{minipage}[b]{0.32\textwidth}
        \centering
        \includegraphics[height=4cm]{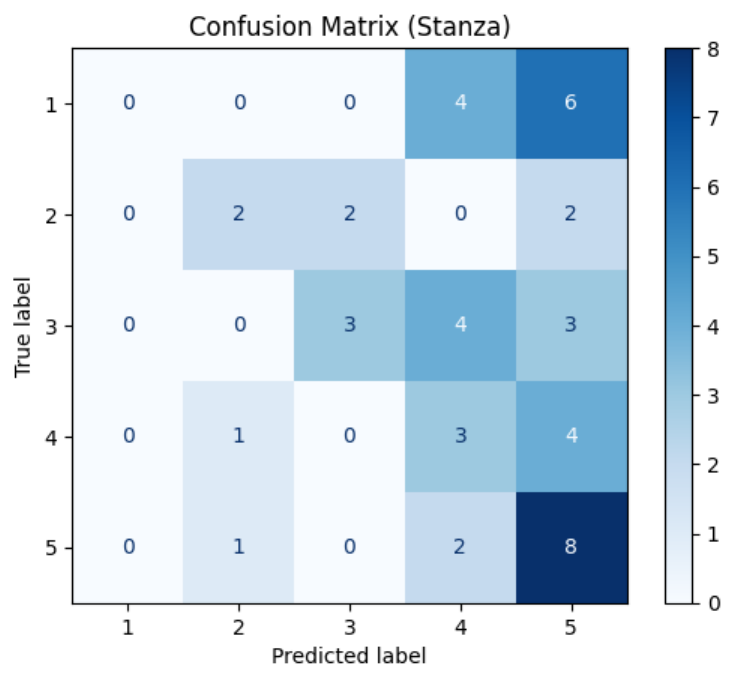}
        \caption*{(d) Stanza}
    \end{minipage}
    \hfill
    \begin{minipage}[b]{0.32\textwidth}
        \centering
        \includegraphics[height=4cm]{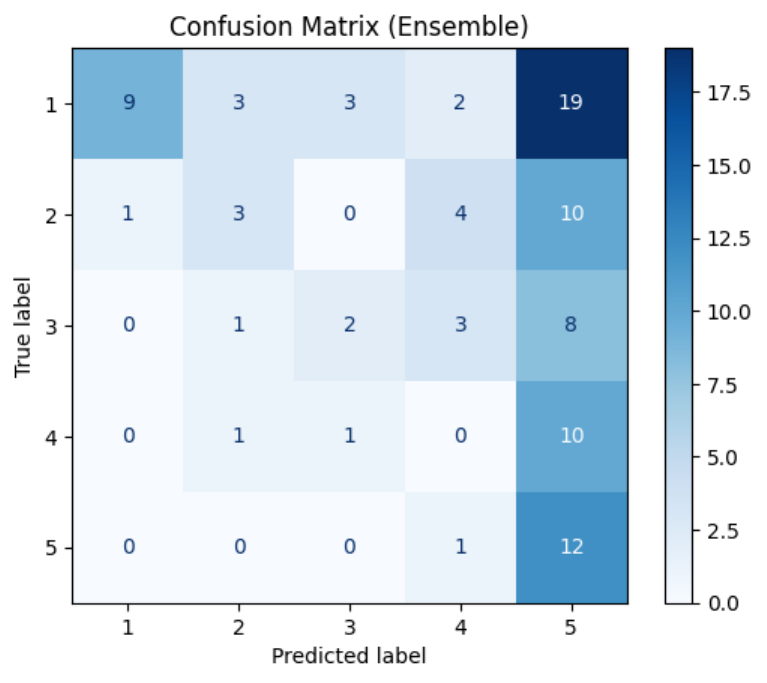}
        \caption*{(e) Ensemble}
    \end{minipage}
    \hfill
    \begin{minipage}[b]{0.32\textwidth}
        \centering
        \includegraphics[height=4cm]{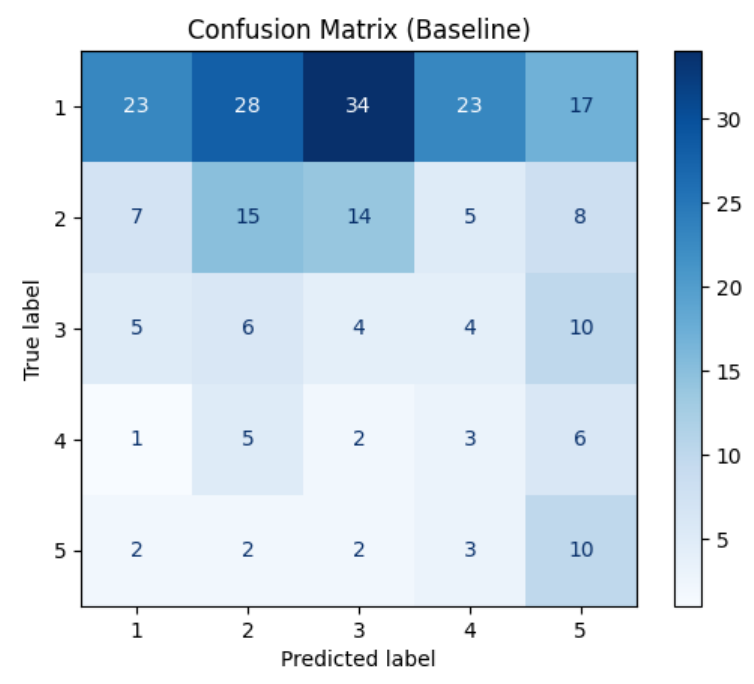}
        \caption*{(f) Baseline}
    \end{minipage}
    \caption{Confusion matrices for all models.}
    \label{fig:cm_all_models}
\end{figure*}

The confusion matrices in Figure~\ref{fig:cm_all_models} reveal a general trend: the models in (a), (b), and (c), which correspond to GPT4o 3-shot, Llama 3-2, and Mistral 7B, tend to struggle with mid-range salience levels (scores 2 and 3), often misclassifying them as higher salience scores such as 4 or 5. In contrast, Stanza (d) and Ensemble (e) models demonstrate a stronger tendency to predict higher salience scores (4 or 5) more frequently. This is likely due to the training data for Stanza and Ensemble being skewed toward entities with higher salience scores, leading these models to overfit on these dominant categories. This observation underscores a potential limitation in the training data distribution, which might favor higher salience entities and influence the predictive bias of these models.

\subsection{Model Performance across Entity Types}
\begin{figure*}[htb]
    \centering
    \begin{minipage}[b]{0.45\textwidth}
        \centering
        \includegraphics[width=\textwidth]{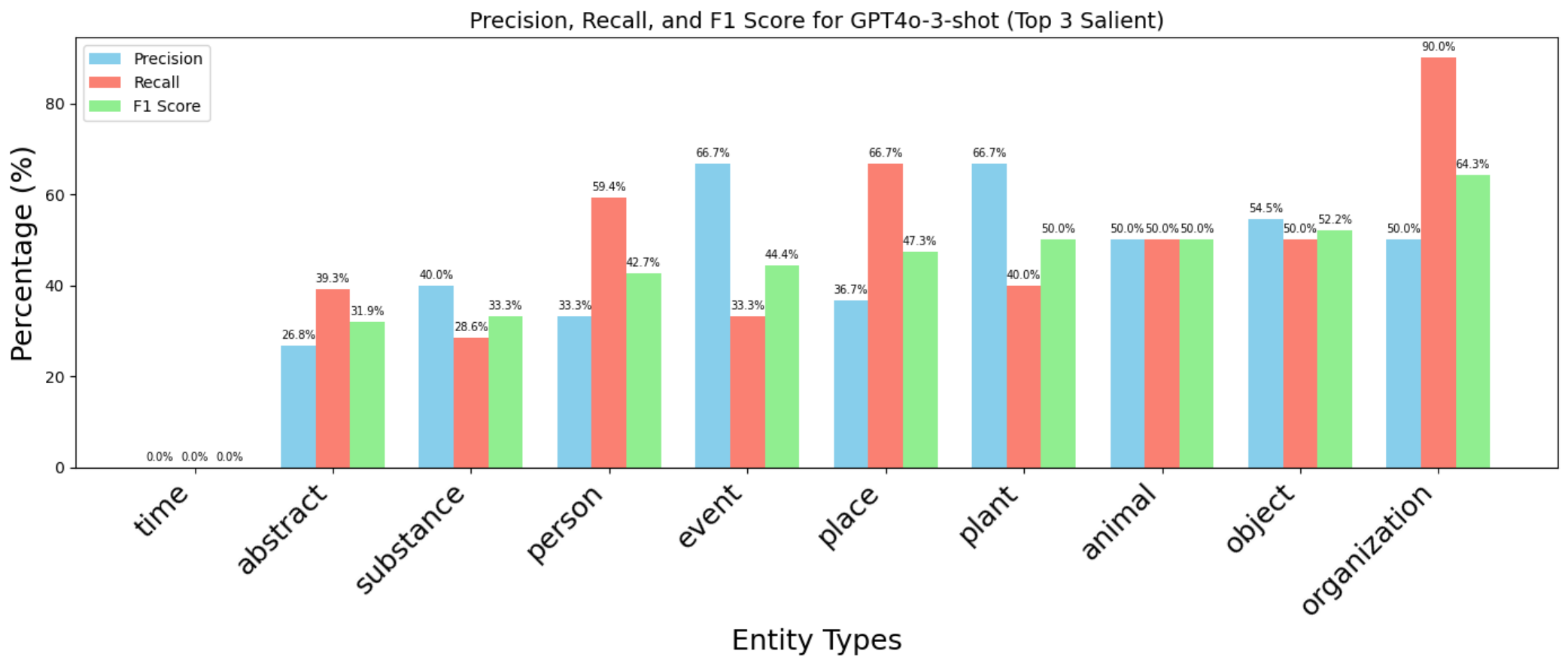}
        \caption*{(a) GPT4o 3-shot}
    \end{minipage}
    \hfill
    \begin{minipage}[b]{0.45\textwidth}
        \centering
        \includegraphics[width=\textwidth]{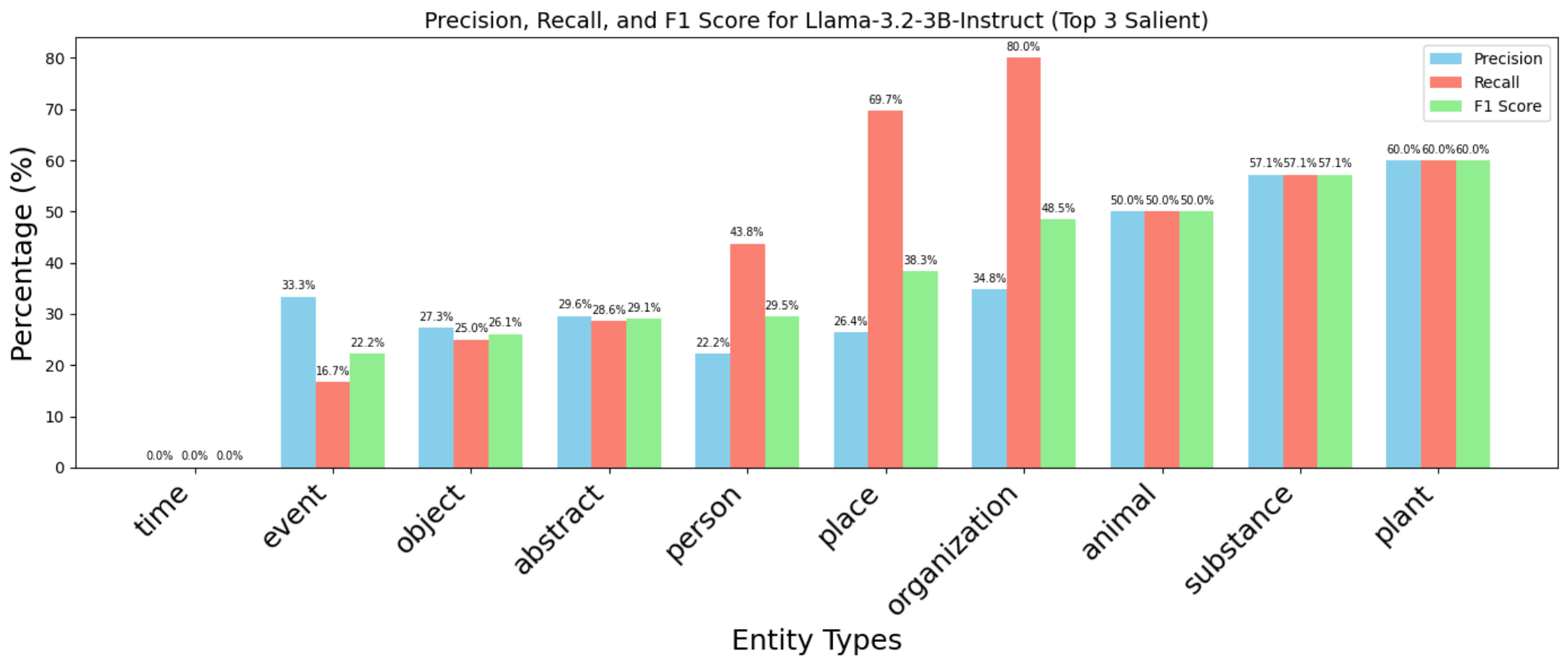}
        \caption*{(b) Llama}
    \end{minipage}
    
    \vspace{0.5cm}  
    \begin{minipage}[t]{\textwidth}
        \begin{minipage}[b]{0.45\textwidth}
            \centering
            \includegraphics[width=\textwidth]{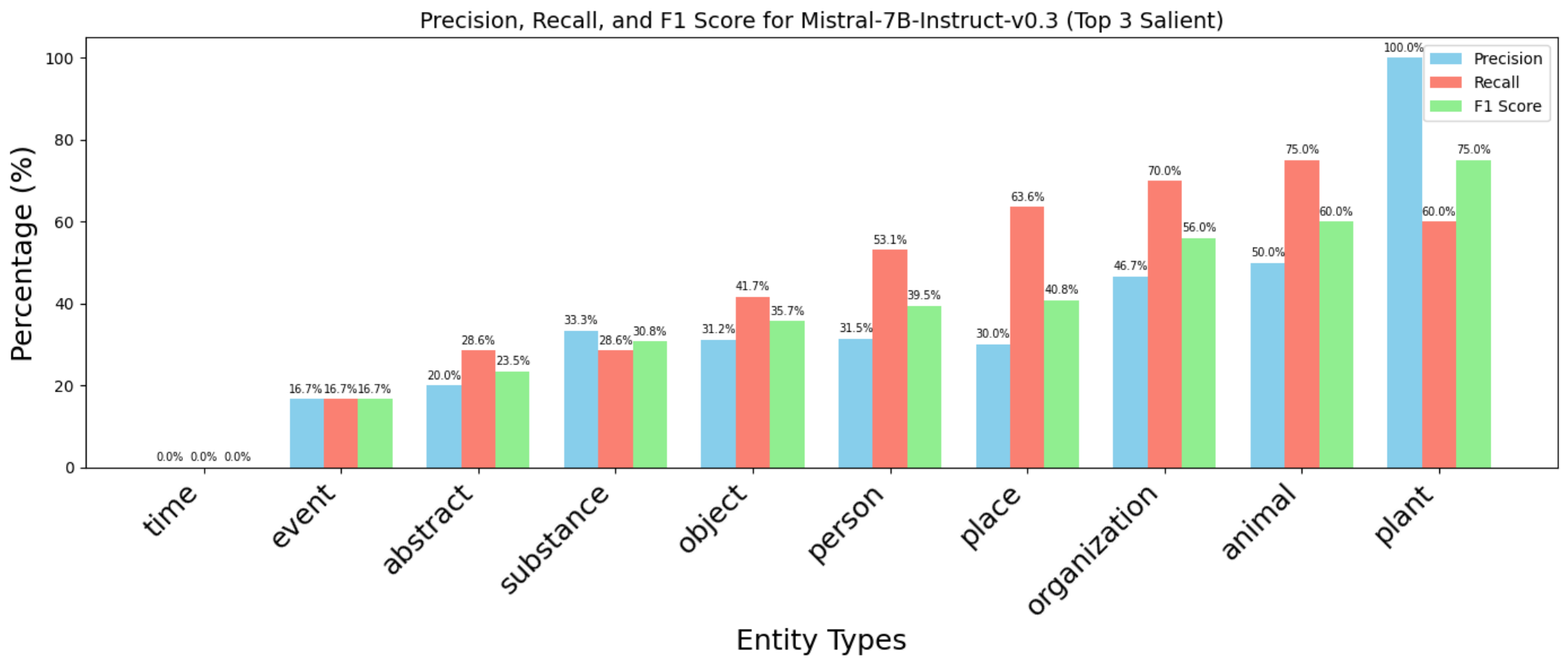}
            \caption*{(c) Mistral}
        \end{minipage}
        \hfill
        \begin{minipage}[b]{0.45\textwidth}
            \centering
            \includegraphics[width=\textwidth]{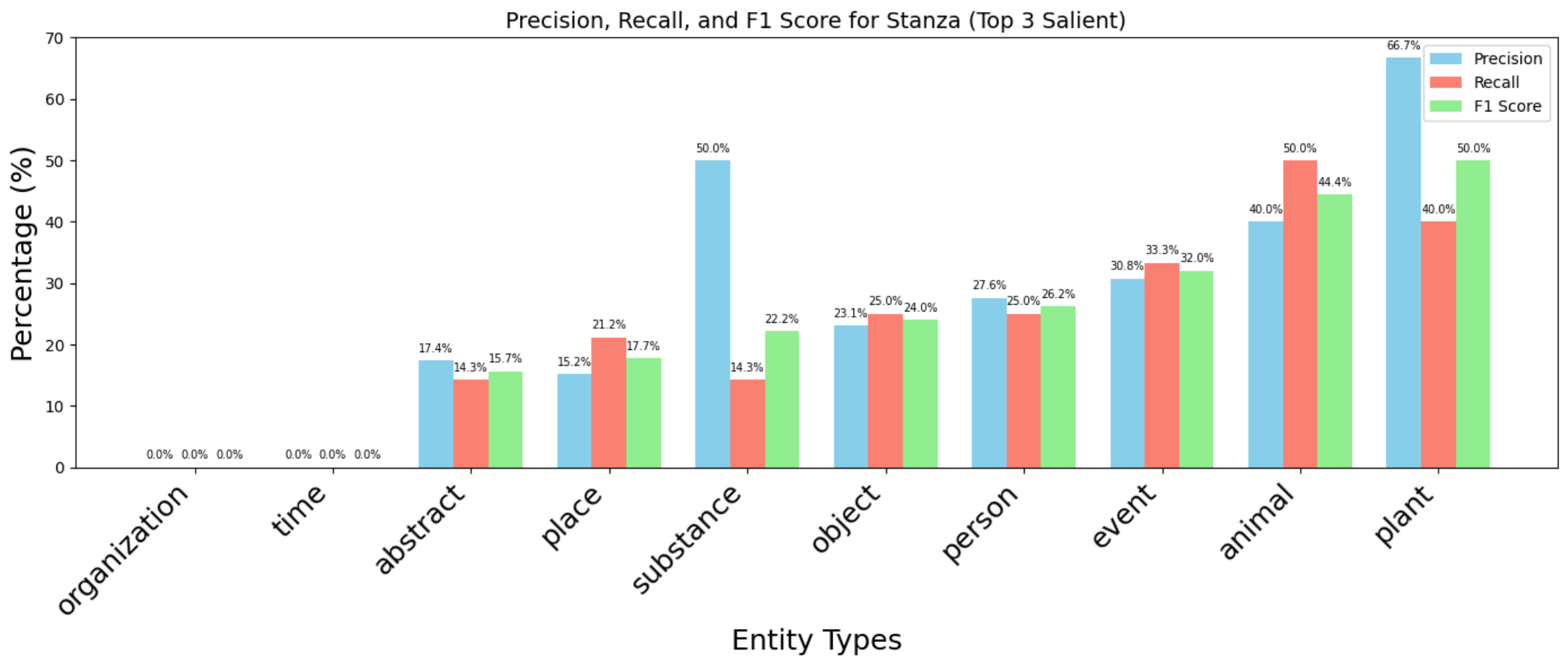}
            \caption*{(d) Stanza}
        \end{minipage}
    \end{minipage}
    
    \vspace{0.5cm}  
    \begin{minipage}[b]{0.45\textwidth}
        \centering
        \includegraphics[width=\textwidth]{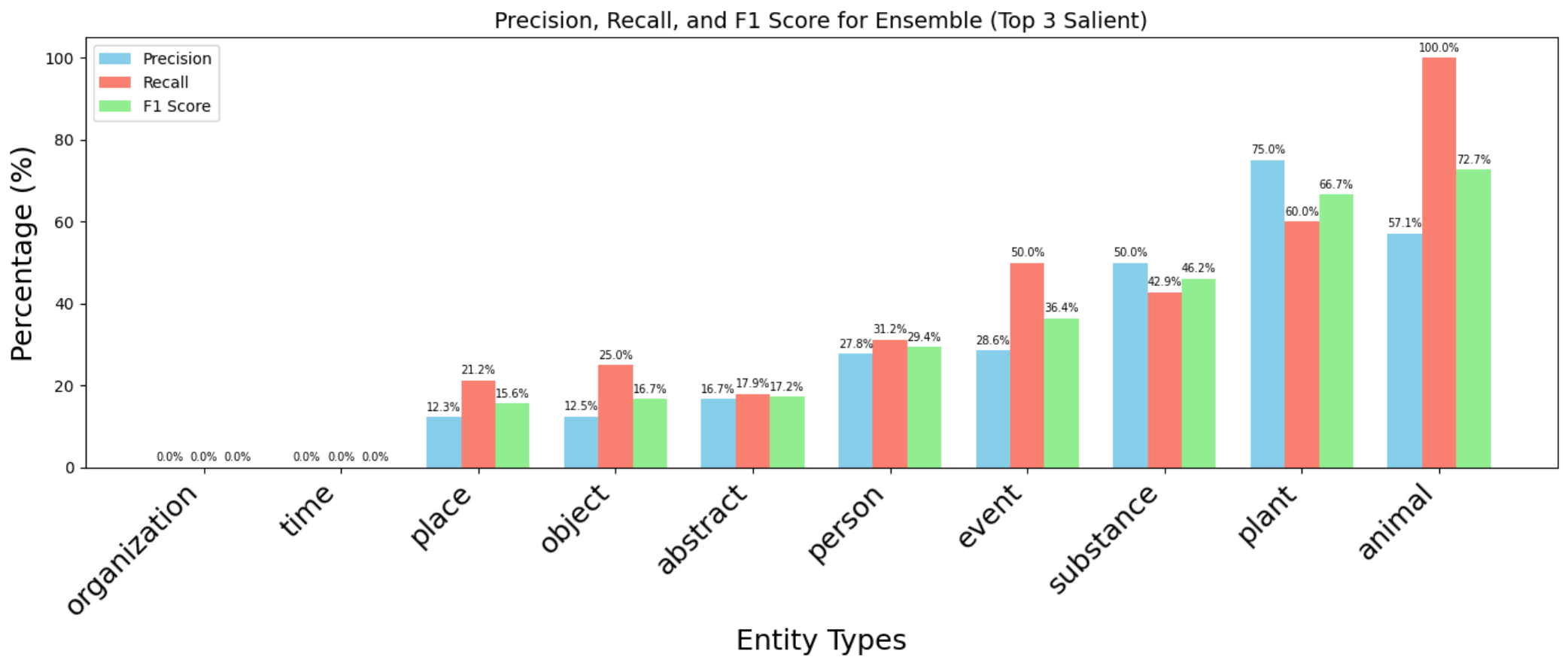}
        \caption*{(e) Ensemble}
    \end{minipage}
    
    \caption{Performance of all models across entity types (Top 3 Salient Entities)}
    \label{fig:prf_all_models}
\end{figure*}

Analysis of model performance across entity types (Figure~\ref{fig:prf_all_models}) reveals several consistent patterns. All models show stronger performance on concrete entities like \textsc{organization} and \textsc{person} compared to abstract or temporal entities. While LLMs (GPT4o and Llama) exhibit higher recall than precision for \textsc{person} entities, suggesting a tendency to overpredict these entity types, both Stanza and Ensemble demonstrate more balanced precision-recall trade-offs across entity types. Most models struggle with \textsc{abstract} entities, showing consistently lower F1 scores for this category compared to other entity types. This pattern suggests that identifying salient abstract concepts remains a key challenge across different architectural approaches.

\begin{figure}[ht]
\centering
\includegraphics [width=0.48\textwidth] {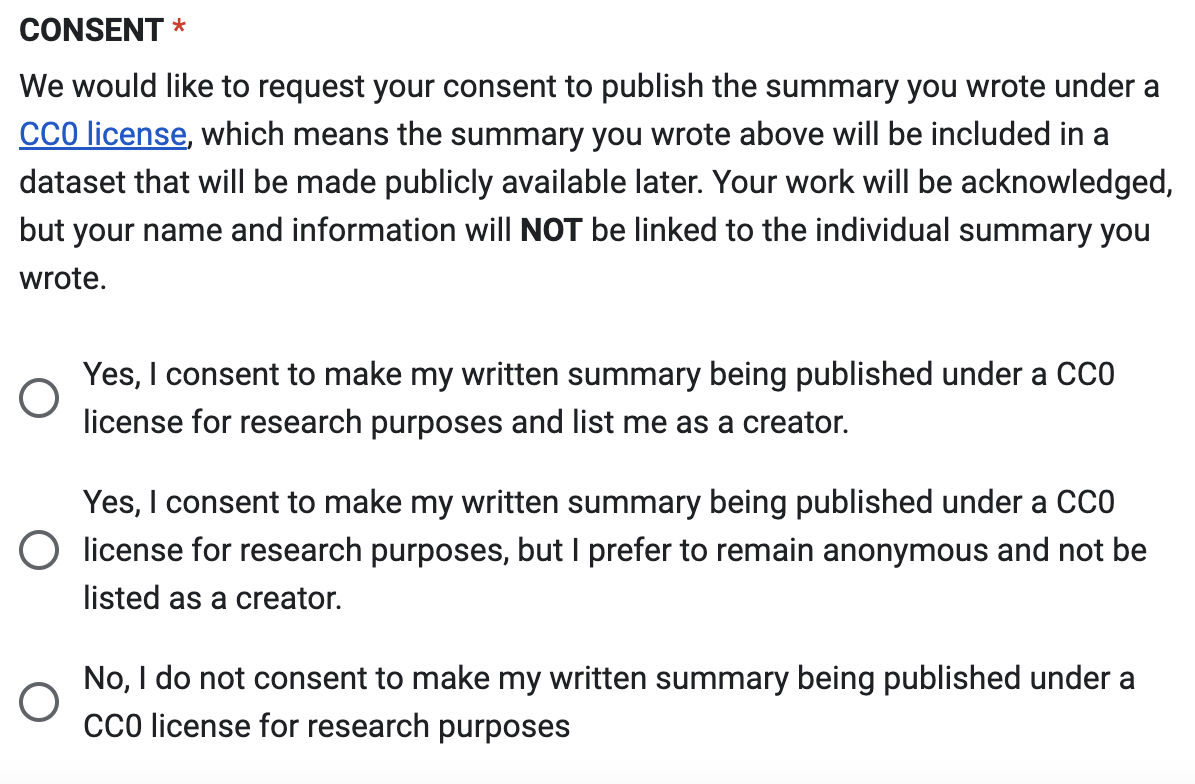}
\caption{A screenshot of the consent form for the annotators.}
\label{fig:AnnoConsent}
\end{figure}

\section{License and Copyright}

All human-written summaries were collected with explicit consent and released under a CC0 license. Annotators were informed that their summaries would be publicly shared for research purposes and were given the option to remain anonymous or receive attribution. All annotators consented to data release. The consent form is shown in Figure~\ref{fig:AnnoConsent}.


\end{document}